# A computable representation of the physical laboratory enables verifiable workflows

Xiaobo Li[1,2,#,*], Luyao Ge[1,#], Xiaohui Li[2], Lulu Guo[1], Ming Mao[3], Jiwang Zheng[3], Wenting Guan[3], Xin Yang[3], Yi Luo[1,2], Jun Jiang[1,2,*], Linjiang Chen[1,2*]

[1] State Key Laboratory of Precision and Intelligent Chemistry, Hefei National Research Center for Physical Sciences at the Microscale, School of Chemistry and Materials Science, University of Science and Technology of China, Hefei, China
[2] Center for Scientific Intelligence Innovation, Hefei, China
[3] Huawei Technologies Co., Ltd.
# These authors contributed equally.
*Correspondence: xiaoboli@ustc.edu.cn; jiangj1@ustc.edu.cn; linjiangchen@ustc.edu.cn.

## Abstract

Making science computable requires representations of both scientific knowledge and the physical world in which scientific claims are tested. A computable representation of the physical laboratory is established through typed research objects, capability-bound operations and a compositional workflow algebra. It provides the physical-world counterpart to machine-readable knowledge, expressing workflows as programs over evolving laboratory states with explicit dependencies, decisions, iteration and concurrency. The representation was implemented in a modular agentic robotic laboratory by binding formal operations to executable Function Skills. For diverse scientific intents, capability-relative workflows were generated, while stateful simulation propagated object transformations and verified operation preconditions and laboratory constraints before dispatch. The proposed representation and its engineering framework jointly establish a general computational interface between agent reasoning and capability-bound physical transformations, providing a foundation for end-to-end autonomous scientific discovery.

## A computable laboratory state–action space

Agentic self-driving laboratories combine artificial-intelligence reasoning with modular systems for synthesis, processing, reaction, characterization and analysis.[1–3] Their central computational problem is not merely to select instruments. Scientific intent is expressed semantically and can vary without bound, whereas execution must conserve samples, respect container and device limits, maintain location, order dependencies and condition later actions on measurements. A plausible plan can therefore be chemically meaningful yet physically inadmissible. Chemical programming languages provide essential operation vocabularies and control structures,[4–6] and laboratory orchestrators coordinate instruments, resources and data.[7–9] Agentic operation therefore requires a computable representation of the physical laboratory, in which research objects, laboratory capabilities and control logic are encoded within a common state-transition semantics.

A chemical procedure links physical operations, evolving samples and measurement-dependent decisions (Fig. 1a). The representation abstracts this procedure as a state–action system for the actionable laboratory—not an idealized chemical universe (Fig. 1b). At workflow step t, St denotes the state of a typed research object, comprising its persistent identity and structured sample, container, evidence and provenance fields. An operation is a typed, workstation-bound function $f_i(\theta_i)$ whose contract declares accepted states, parameters, constraints, effects, returns and failures. When its mandatory predicates are satisfied, the operation maps $S_t$ to $S_{t+1}$ and records evidence; otherwise, it returns a diagnostic without changing the affected simulated state. The computational unit is thus a physically bounded state transition, rather than a free-text instruction or an isolated device call (Supplementary Notes 1–2, Supplementary Tables S1–S3, Supplementary Fig. S1 and Supplementary Data 1).

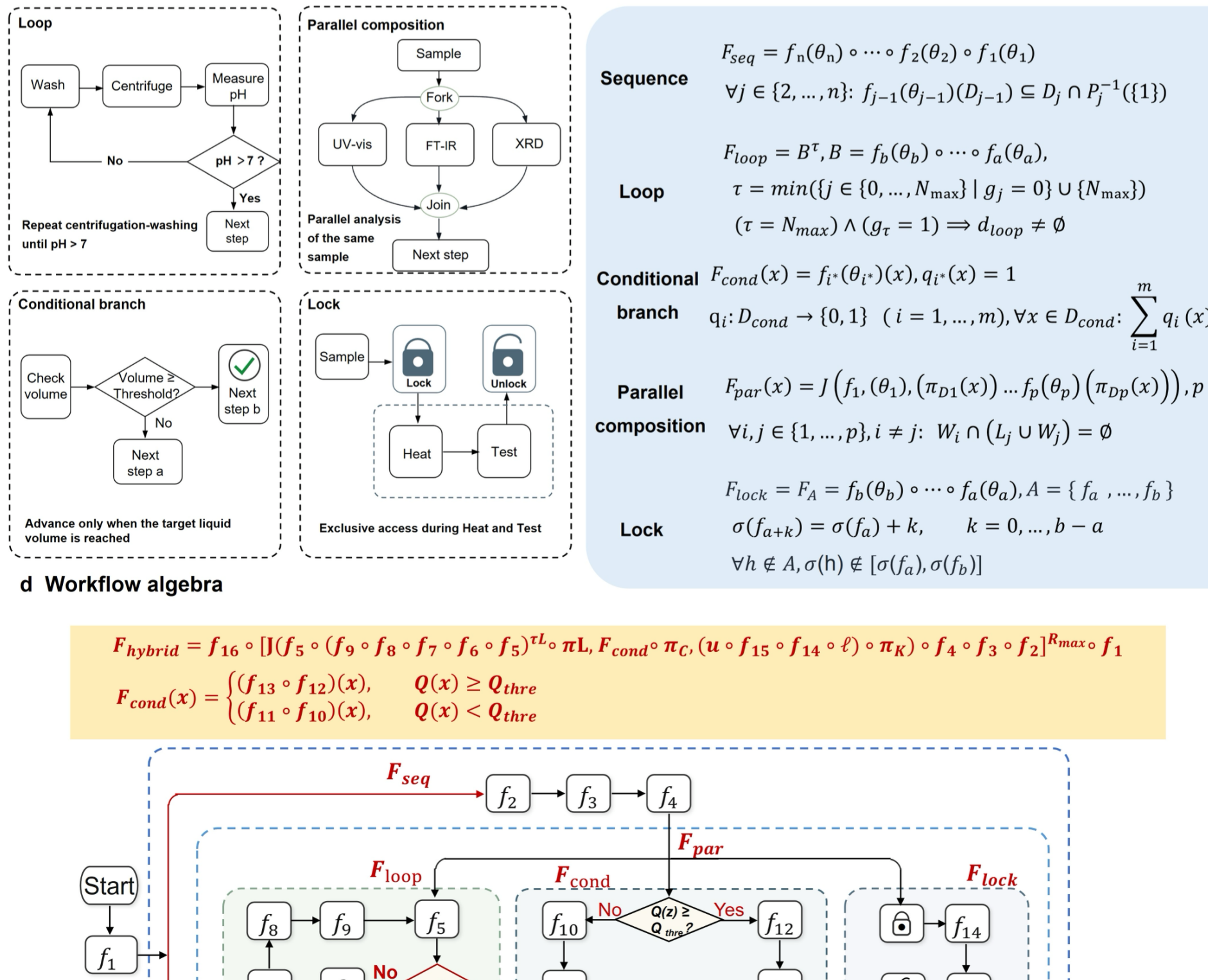

**Fig. 1 |** A computable representation of the physical laboratory. a, A chemical procedure comprises transformations of samples and containers, with measurements and control decisions determining subsequent actions. b, The same procedure is represented as a state–action system: St denotes a typed research-object state, and each capability-bound operation fi(θi) declares accepted inputs, constraints, effects, returns and failures before mapping $S_t$ to $S_{t+1}$. c, Five operators encode sequence, bounded iteration, evidence-dependent branching, parallel fork–join and protected execution. d, Recursive nesting of operations and operators forms a workflow algebra over evolving laboratory state. The panels therefore progress from physical procedure to computable state, explicit control logic and compositional program. XRD, X-ray diffraction; UV–vis, ultraviolet–visible spectroscopy.

## Workflows as compositional programs

Primitive transitions become workflows through five operators (Fig. 1c). Sequence propagates state between operations. A bounded loop makes iteration and non-convergence explicit. A conditional selects a branch from current evidence. Parallel fork–join projects objects into independent branches and reconciles compatible updates. A protected interval keeps a resource-sensitive block contiguous. The workflow algebra nests these operators (Fig. 1d), allowing a scientific procedure to be expressed as a program over evolving laboratory state. Type compatibility is checked at sequential boundaries; loops require finite bounds; conditional guards must resolve to a permitted branch; parallel branches must not introduce conflicting writes; and protected blocks preserve declared ordering. Workflow abstraction therefore converts experimental logic into structures that can be composed and computationally inspected (Supplementary Notes 3–4, Supplementary Table S4, Supplementary Fig. S2 and Supplementary Data 2).

This construction separates scientific plausibility from operational admissibility. The formal layer tests the latter against an explicit capability inventory and cannot invent unavailable operations. Its generative space is the closure of registered primitives under the five operators. Generality is capability-relative: transfer requires each laboratory to register local operations, constraints and adapters.

## Engineering executable semantics

To test whether the representation can be operationalized, it was implemented in a modular agentic self-driving laboratory (Fig. 2).[10] Forty-nine workstation specifications—45 physical and 4 virtual control-flow modules—were converted from structured device descriptions into agent-readable Function Skills. The frozen snapshot exposes 63 functions covering liquid and solid handling, mixing, heating, separation, reaction, electrochemical testing, chromatography, spectroscopy, diffraction, imaging and control flow. Each Function Skill preserves the corresponding operation contract and uses workstation-qualified names where repeated actions could otherwise be ambiguous. This compilation turns the installed laboratory into a machine-readable capability space, while keeping the formal representation independent of any single device command set.

Scientific intent is translated by an agent into a Python workflow containing only registered calls. Before dispatch, the abstract syntax tree is admitted against allow-listed imports, functions and entry points, then normalized conservatively. The resulting program enters a shared stateful simulator. Each call evaluates all encoded mandatory and advisory conditions before applying effects. A mandatory failure blocks the update; a successful call commits the new object state and appends a step record. Diagnostics localize the function, step, observed condition, expected condition and suggested remediation. This validate-then-commit cycle allows a workflow fragment to be revised without treating the entire protocol as unstructured text. It verifies operational consistency with encoded contracts; it does not prove chemical truth, experimental safety beyond those rules or device reliability (Supplementary Notes 5–7, Supplementary Tables S5–S8, Supplementary Figs. S3–S6, Supplementary Data 1 and 4, and Additional Files 1–2).

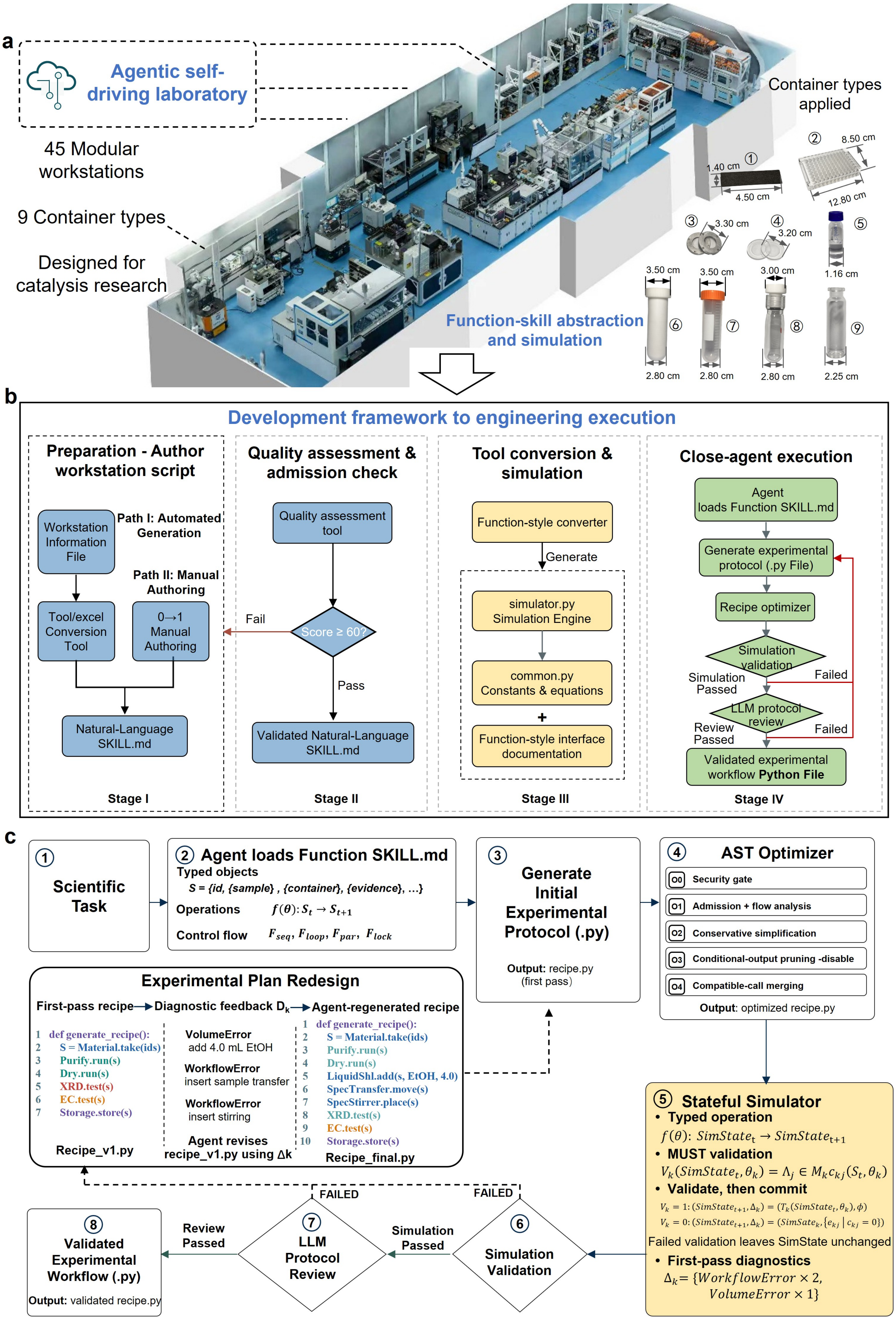

**Fig. 2 |** Engineering and checking computable laboratory workflows. a, Structured specifications register the physical and virtual capabilities of the modular agentic self-driving laboratory. b, These records are compiled into agent-readable Function Skills, shared runtime types and simulator functions while preserving operation contracts. c, Scientific intent is compiled into a Python workflow containing registered calls, admitted through abstract-syntax-tree analysis and evaluated by stateful simulation before review and downstream serialization. Successful calls commit object-state changes; failed MUST predicates leave $SimState_t$ unchanged and return a localized diagnostic set $\Delta_k$ for regeneration. Here, $S_t$ denotes a research-object state, and $SimState_t$ denotes the bounded executable projection used by the simulator. O0–O4 denote optimizer levels and are unrelated to these state symbols.

## Capability-relative generation from scientific intent

Compositional scope was examined with four prompts that specified scientific objectives and control logic but not complete workstation sequences (Extended Data Fig. 1). The first compared five Fe-introduction routes for NiFe-based catalysts under matched alkaline oxygen-evolution testing and produced a 27-node sequence across 11 workstations. The second coupled layered double hydroxide synthesis and electrode preparation in a protected interval to HMF oxidation, then forked solid-electrode and liquid-product objects into three analytical branches; it contained 22 physical nodes, 31 simulated steps and 11 workstations. The third combined furfural hydrogenation with a bounded gas-chromatography method-optimization loop ($N_{max}$ = 6), yielding 20 nodes and 23 simulated steps. The fourth coupled ZnO preparation to parallel infrared and ultraviolet–visible characterization, with a conditional dilution loop ($N_{max}$ = 3), yielding 27 nodes, 30 simulated steps and 15 workstations.

All four accepted workflows used only registered capabilities and passed structural analysis and state simulation with no mandatory errors or advisory warnings. The cases span sequence, conditional branching, bounded iteration, fork–join parallelism and protected execution. They are not four hard-coded protocol templates; each is a different composition of shared operations selected through required object transitions and evidence dependencies (Supplementary Notes 8–11, Supplementary Tables S9–S11, Supplementary Figs. S7–S10, Supplementary Data 3 and Additional File 3).

## A representation layer for agentic laboratories

Recent efforts have added a foundational laboratory schema, literature-to-protocol agents and simulation-guided refinement to the established programming and orchestration layers.[11–13] Broader developments encompass modular self-driving-laboratory architectures, autonomous materials synthesis and distributed closed-loop discovery,[14–17] while knowledge graphs, natural-language robotic planning, orchestration middleware and sample-centred automata address complementary parts of laboratory autonomy.[18–21] The contribution here is the unification of three levels that are commonly separated: a typed representation of physical research objects, an algebra for composing capability-bound transitions, and an executable realization that checks those transitions against evolving state. Unlike protocol-centric representations, the object state being transformed remains an explicit operand throughout generation, composition and checking. Agent reasoning is thereby connected to physical action through an explicit computational object rather than through prose alone.

The present checks are intentionally bounded. Their validity depends on the completeness of workstation contracts and the fidelity of simulated state. Scheduler-level resource isolation, live telemetry, device faults, empirical repeatability and chemical outcomes require execution evidence or additional models. These are extension interfaces rather than hidden assumptions: constraints, resources, observations and adapters can be added without changing the intent layer. Local failures can consequently be assigned to an object, operation, guard or join rather than to the workflow as a whole. By making laboratory state, permitted transformations and control structure jointly computable, the framework establishes a basis for generating increasingly complex experiments while keeping their physical commitments verifiable before execution.

## Methods

### Research-object and operation representation

At workflow step $t$, $S_t$ denotes the state of a typed research object. Each object contains a persistent identifier and typed subrecords for sample identity and composition, phase, amount, container type and state, location, evidence and provenance. Evidence may contain scalar observations or references to generated data. A workstation contract declares the projection of these fields required for an operation. The current simulator implements the subset needed for workstation checking: container identifier and type, closure state, sample state, volume, mass, location, source-bottle use, step logs, errors and warnings.

Each physical operation is represented by a Function Skill containing the workstation binding, operation name, input-state requirements, cautions, typed parameters, constraints, state changes, return schema and diagnostic categories. For operation $f_i$ with parameters $\theta_i$, a transition from $S_t$ to $S_{t+1}$ is admitted only when all mandatory predicates $P_i(S_t, \theta_i)$ hold. For operations involving multiple research objects, the function acts on the corresponding tuple of object states. Advisory predicates record warnings but do not block the transition. Mandatory failure returns an unsuccessful result and leaves the affected simulated state unchanged.

### Workflow composition

Sequence was defined by functional composition, with each intermediate output required to lie in the next operation's input domain. A loop repeated its body until the continuation guard became false or a declared $N_{max}$ was reached; an active guard at the bound returned a non-convergence diagnostic. Conditional composition required a unique permitted branch. Parallel fork–join projected the objects required by each branch and admitted the join only when branch write sets did not conflict and shared updates agreed. A protected interval was constrained to occupy consecutive scheduler positions. The five operators could be nested recursively.

### Capability compilation

Workstation capabilities were supplied as structured records or curated Skill documents. Required fields included workstation identity, available operations, accepted input and output states, parameters, bounds, container requirements, state effects and returns. A development-stage quality gate assessed documentation completeness; the archived configuration admitted records at the required-field score of 60. Accepted records were converted into workstation-qualified interfaces, shared data classes and simulator implementations. The frozen inventory contains 28 synthesis, 5 reaction-and-testing, 12 characterization and 4 virtual specifications, declaring 63 functions.

### Structural analysis and stateful simulation

Generated workflows were Python modules with one generate_recipe() entry point. Each module was parsed into an abstract syntax tree. Admission permitted only registered runtime imports and allow-listed function calls, and rejected wildcard imports, dynamic calls, unsupported top-level statements, forbidden names and additional function definitions. Literal arithmetic and Boolean expressions were folded, inactive constant branches were removed and redundant pass statements were eliminated. Conditional-output pruning was available as a domain-aware transformation but was disabled in the frozen configuration.

For simulation, the shared state was reset, artificial delays were disabled and registered workstation functions were injected into the workflow namespace. Each operation collected all applicable checks before applying any effect. Checks covered container existence and count, container and sample state, cumulative and per-call volume, mass, parameter ranges, file constraints and declared workstation dependencies. A diagnostic stored the function, step, code, category, expected and observed values, parameter snapshot, severity and suggested remediation. Mandatory diagnostics blocked passage. Successful operations committed state changes and appended a structured step record.

### Case construction and evaluation

The four prompts requested Fe coordination-environment modulation and matched alkaline oxygen-evolution testing; layered double hydroxide synthesis, electrochemical HMF oxidation and parallel analysis; furfural hydrogenation with iterative gas-chromatography optimization; and ZnO synthesis with parallel spectroscopy and adaptive dilution. Complete workstation sequences were not supplied. The agent received the controller Skill and registered Function Skills. The workflow-generation agent was powered by GPT-5.6 Sol (OpenAI). Model invocations used in this study were performed between 15 and 25 August 2026. Accepted workflows were serialized into canonical JSON, and node counts, workstation counts, simulated steps and diagnostic totals were calculated from the frozen workflow and validation records. No workflow was physically executed for this study.

### Statistical analysis

The study evaluates a representation, its software realization and four workflow constructions. No wet-laboratory outcome comparison or inferential statistical analysis was performed.

## Data availability

The typed capability registry, formal definitions, four prompts, canonical workflow representations, validation summaries and evidence manifest are provided as Supplementary Data 1–4. Complete workflow and validation records for the four cases shown in Extended Data Fig. 1 are supplied as Additional File 3. No new wet-laboratory measurement data were generated for the workflows reported here.

## Code availability

Source code for Function-Skill conversion, workflow analysis, simulation and serialization is supplied as Additional File 1, and the frozen Function-Skill library as Additional File 2. Additional File 4 contains the internal release-page record supplied for provenance only; it is not presented as a public repository, persistent public access route or DOI.

## Acknowledgements

The workflow generation and stateful simulations were performed on the Robotic AI-Scientist Platform of the Chinese Academy of Sciences (CAS).

## Author contributions

X.L. and L.G. contributed equally to this work. X.L., L.C. and J.J. conceived and supervised the project. X.L., L.G., X.H.L., L.G., M.M., J.Z., W.G. and X.Y. developed the formal representation and engineering framework, constructed and analysed the workflows, and curated the supporting materials. All authors discussed the results, revised the manuscript and approved the final version.

## Competing interests

The authors declare no competing interests.

## Additional information

Supplementary information accompanies this paper. Reproducibility settings and the software and evidence manifest are provided in Supplementary Note 12, Supplementary Table S12 and Supplementary Data 4; the corresponding source-code archive, Function-Skill library, workflow archive and internal release-page record are supplied as Additional Files 1–4.

**Prompt 1:**
**Design an experiment to investigate how the existing form and coordination environment of Fe modulate the alkaline oxygen evolution reaction (OER) performance of NiFe-based catalysts. NiFe-based catalytic materials with different Fe introduction methods and distinct Fe coordination environments are required to be fabricated.**

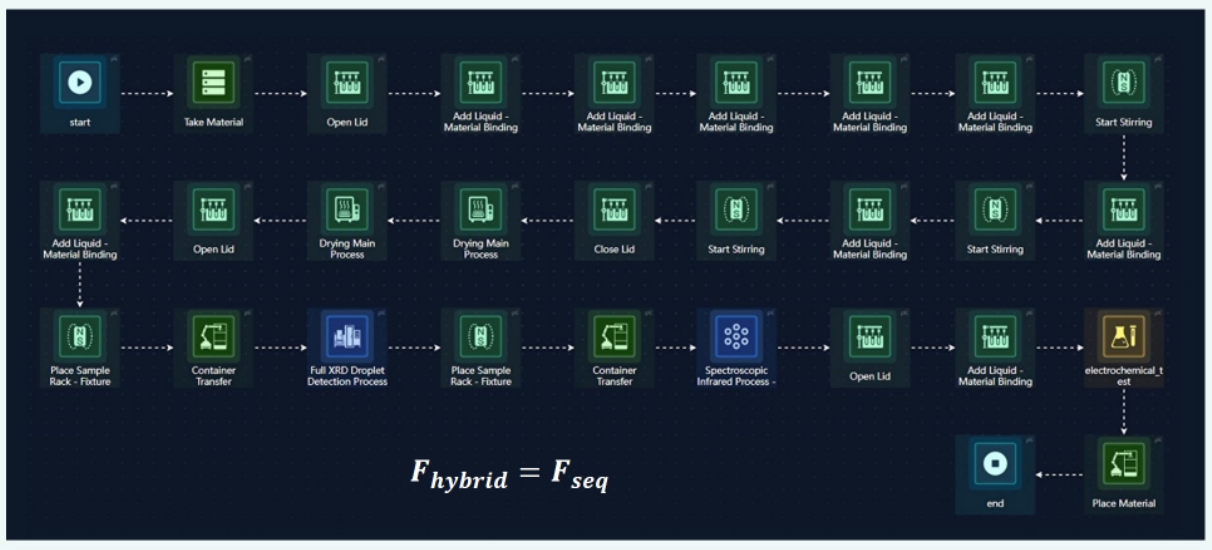

**Prompt 2:**
**Design an integrated synthesis, electrochemical testing and multimodal characterization experiment for LDH catalysts toward alkaline 5-hydroxymethylfurfural (HMF) oxidation. The catalyst preparation and electrode-fabrication steps must follow a time-locked sequential workflow. After electrochemical testing, the solid electrode and liquid reaction products are required to undergo parallel contact-angle measurement, XRD analysis and liquid-phase product analysis, followed by integrated evaluation of the results.**

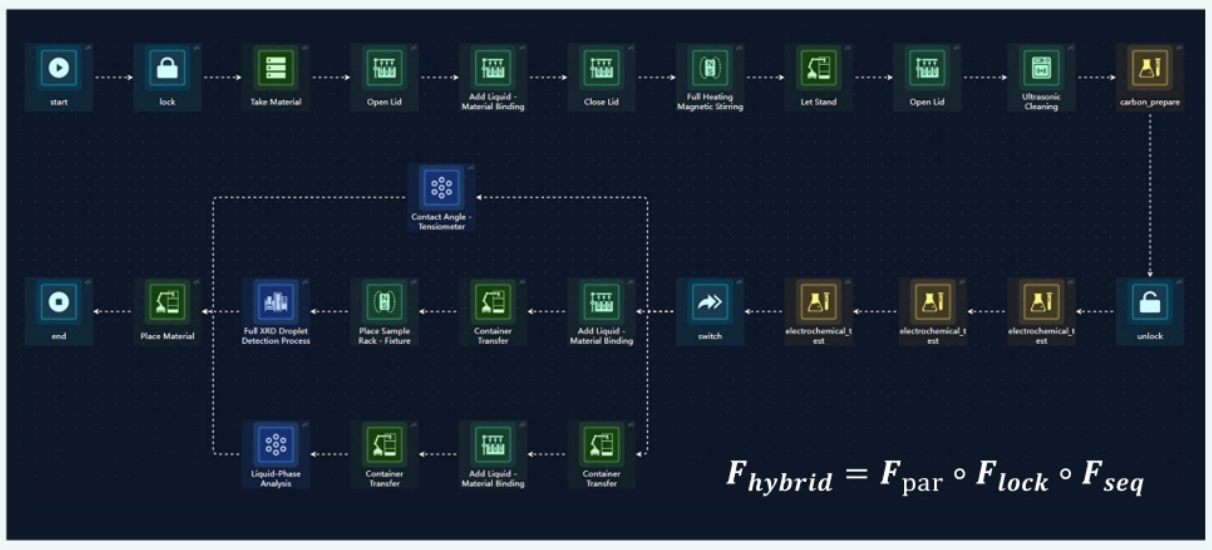

**Prompt 3:**
**Design an automated catalyst synthesis and closed-loop analytical optimization experiment for the electrocatalytic hydrogenation of furfural to furfuryl alcohol. After catalyst preparation, a closed-loop algorithm is required to iteratively optimize the gas-chromatographic separation of furfural, furfuryl alcohol and major liquid by-products until predefined peak-resolution criteria are satisfied. The optimized method will then be applied to quantify the reaction mixture and evaluate furfural conversion, furfuryl alcohol yield and product selectivity. Need to use a loop node.**

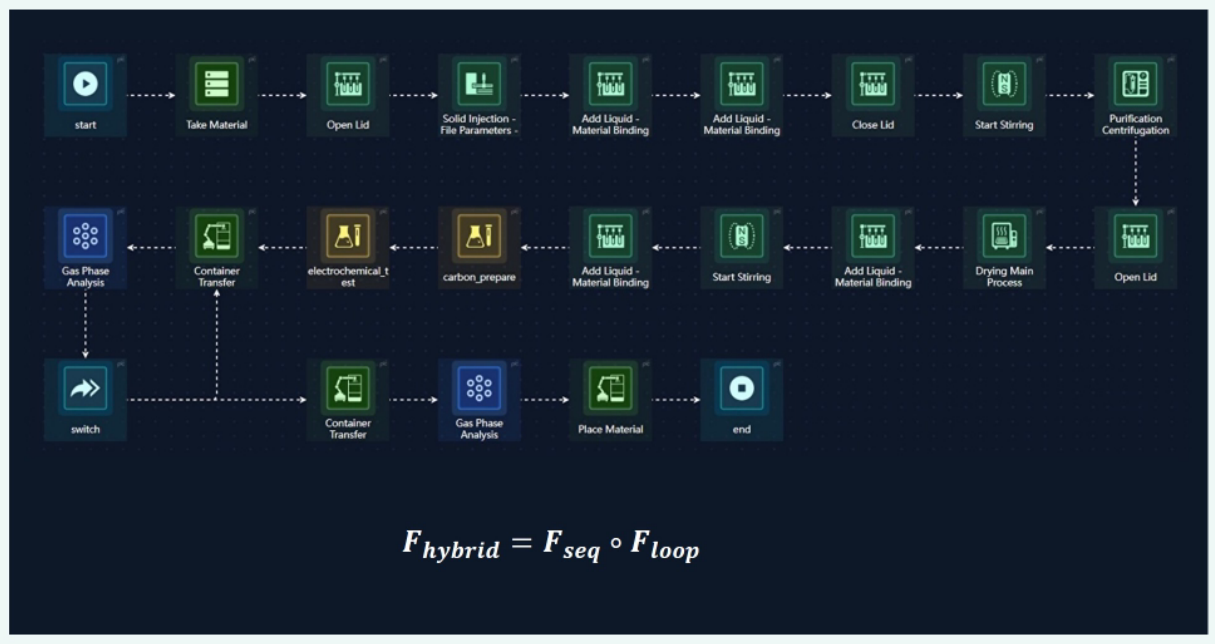

**Prompt 4:**
**Design an automated synthesis and adaptive spectroscopic characterization experiment for ZnO nanoparticles. Following synthesis, centrifugal purification, drying and redispersion, the samples are required to undergo parallel FTIR and UV–vis spectroscopic characterization. In the UV–vis branch, a conditional dilution-and-remeasurement loop must be implemented until the measured absorbance falls within the predefined quantitative range.**

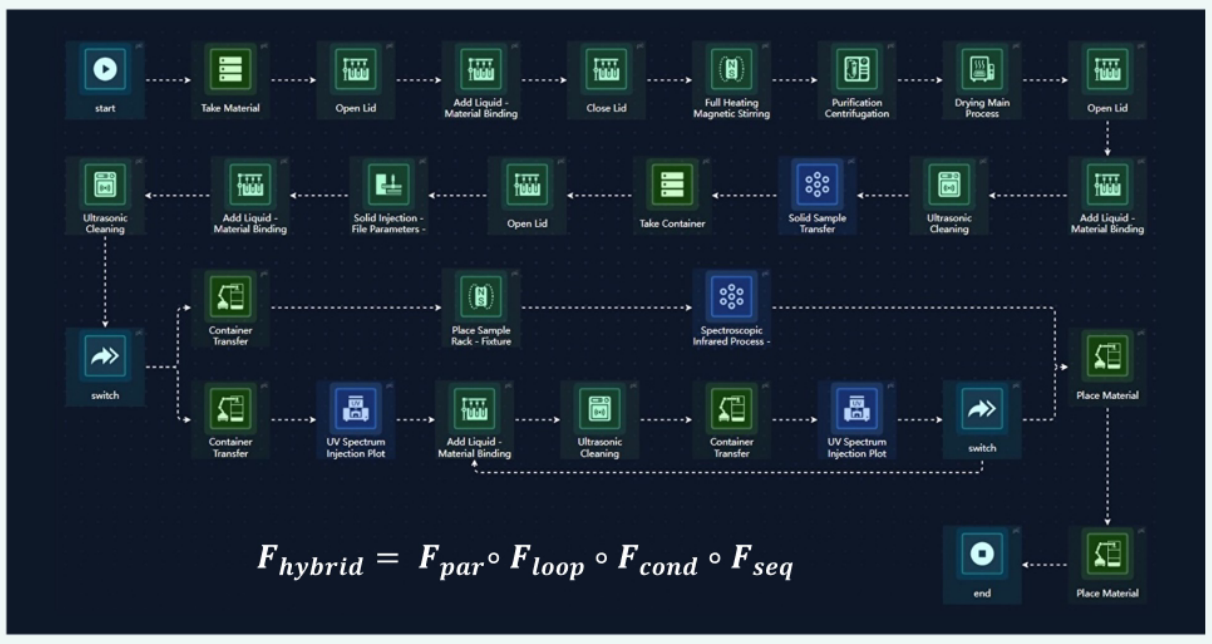

**Extended Data Fig. 1 |** Capability-relative workflows generated from four scientific intents. Each panel pairs a prompt with its generated workflow: a, five routes to modulate the Fe coordination environment of NiFe-based catalysts followed by matched alkaline oxygen-evolution testing; b, layered double hydroxide synthesis and electrode preparation within a protected interval, electrochemical HMF oxidation, and fork–join analysis of solid-electrode and liquid-product objects; c, furfural hydrogenation coupled to a bounded gas-chromatography method-optimization loop; and d, ZnO synthesis followed by parallel infrared and ultraviolet–visible spectroscopy and a bounded evidence-triggered dilution loop. The graphs expose ordering, branching, iteration, synchronization and protected execution composed from registered capabilities. They report pre-dispatch workflow generation and verification by structural analysis and stateful simulation, not wet-laboratory outcomes. HMF, 5-hydroxymethylfurfural.

# Supplementary Information

## A computable representation of the physical laboratory enables verifiable workflows

Xiaobo Li[1,2,#,*], Luyao Ge[1,#], Xiaohui Li[2], Lulu Guo[1], Ming Mao[3], Jiwang Zheng[3], Wenting Guan[3], Xin Yang[3], Yi Luo[1,2], Jun Jiang[1,2,*], Linjiang Chen[1,2*]

[1] State Key Laboratory of Precision and Intelligent Chemistry, Hefei National Research Center for Physical Sciences at the Microscale, School of Chemistry and Materials Science, University of Science and Technology of China, Hefei, China
[2] Center for Scientific Intelligence Innovation, Hefei, China
[3] Huawei Technologies Co., Ltd.
# These authors contributed equally.
*Correspondence: xiaoboli@ustc.edu.cn; jiangj1@ustc.edu.cn; linjiangchen@ustc.edu.cn.

This Supporting Information contains Supplementary Notes 1–12, Supplementary Tables S1–S12 and Supplementary Figs. S1–S10. Supplementary Data files and author-supplied Additional Files are described in the SI Guide at the end of this document. References cited here follow the numbering of the main Article.

### Supplementary Notes

#### Supplementary Note 1. Typed research-object representation

The workflow representation begins with a laboratory state, denoted by $S_t$ at workflow step t, comprising one or more typed research objects. Fig. 1a,b defines the object schema as

S={id,{sample},{container},{evidence},…}.

The representation binds scientific material to the physical context required for manipulation. The sample component records identity, composition, phase, mass and volume. The container component records type, geometry, capacity, lid or holder state and other handling-relevant properties. The evidence component records observations and data products, including temperature, pH, colour and characterization outputs. The ellipsis permits laboratory-specific fields without altering the required object–container–evidence structure.

Fig. 1a,b illustrates the representation through three research-object states. $S_1$ contains a suspension of Chemicals A and B in a lidded 50 ml tube with no evidence recorded. $S_5$

contains a liquid–solid object in an open 50 ml tube with pH evidence. $S_{14}$ contains a dry solid derived from $S_{13}$, placed in a characterization holder with XRD and UV–vis evidence. These examples show that provenance is retained while the material, container and evidence states evolve.

The executable simulator implements a bounded projection of this conceptual object. The supplied $SimState_t$ records containers, errors, warnings, step logs, source-bottle usage and a step counter. Each Container records ***container_id***, ***container_type***, ***container_status***, ***sample_status***, ***sample_volume_ml***, ***sample_mass_g*** and ***location***. This implementation does not replace the richer definition of St. It provides the subset required to test the currently encoded workstation contracts. The conceptual and implemented fields are aligned in Supplementary Table S2.

Research objects are identified independently from workflow operations. An operation can therefore accept one or more objects, test their current states, update approved fields and append evidence without discarding object identity. This separation is required for provenance, branching and convergence. It also permits two branches to operate on distinct projections of the same parent object and later join their evidence, provided that the parallel consistency conditions in Supplementary Note 3 are satisfied.

**Cross-references: Fig. 1a,b; Supplementary Tables S1 and S2; Supplementary Fig. S1; Supplementary Data 1.**

**Supplementary Note 2. Primitive operation contracts**

A primitive experimental operation is represented by the state transition

$$\boldsymbol{f_i(\theta_i): S_t \rightarrow S_{t+1},\ \ P_i(S_t, \theta_i) = 1} \quad (1)$$

where $f_i$ is operation $i$, $\theta_i$ contains its arguments and $P_i$ is its precondition predicate. The operation is admissible only when the predicate evaluates to one. Its output must remain in the domain accepted by the next operation.

The Function-Skill interface realizes this primitive contract as a workstation-bound function declaration. The supplied documents define the interface through workstation identity, preconditions, cautions, typed arguments, constraints, state changes, structured returns and declared failure categories. Each physical function uses a namespace call, for example ***Liquid_Handling_Station_1ml_V2.open_caps(...)***. The namespace encodes workstation ownership and removes ambiguity when different devices expose similarly named operations.

Function signatures use explicit types and unit-bearing parameter names. Enumerations are constrained through Literal aliases, nested inputs are represented by named data classes, and shared physical limits are centralized in ***_common.py***. A Function-Skill document contains only the callable interface and contract. The corresponding implementation in ***simulator.py*** performs the actual parameter, precondition and state-transition checks. The interface and implementation are required to remain synchronized.

Constraints are separated into MUST and SHOULD rules. A failed MUST rule is added to SimState.errors and blocks acceptance. A failed SHOULD rule is added to ***SimState.warnings*** and does not block acceptance. State changes are committed only when every MUST rule for the operation passes. Data-producing operations return typed result objects containing file or measurement references in addition to the common ***StepResult*** fields.

The current common runtime enumerates 11 container types: sample vial, 50-ml heat-resistant bottle, screw vial, chromatography vial, carbon-paper holder, 10-ml pressure reaction tube, test holder, 96-well plastic plate (96-well quartz plate) and 54-position GC plate (54-position LC plate). Laboratory-specific limits remain attached to the corresponding Function-Skill and shared constant, rather than being generalized beyond the documented capability.

**Cross-references: Fig. 1b; Supplementary Table S3; Supplementary Data 1.**

## Supplementary Note 3. Mathematical semantics of workflow composition

The five control-flow operators in Fig. 1c organize primitive transitions without changing their individual contracts. Equations (2)–(14) below reproduce the supplied mathematical-definition document. Symbols are defined in Supplementary Table S1.

### Sequential composition

$$\boldsymbol{F_{seq}} = \boldsymbol{f_n}(\boldsymbol{\theta_n}) \circ \cdots \circ \boldsymbol{f}_2(\boldsymbol{\theta}_2) \circ \boldsymbol{f}_1(\boldsymbol{\theta}_1) \tag{2}$$

Execution proceeds from right to left. Every intermediate state must satisfy the next precondition and fall within the next domain:

$$\forall \boldsymbol{j} \in \{\boldsymbol{2}, \dots, \boldsymbol{n}\}: \quad \boldsymbol{f_{j-1}}(\boldsymbol{\theta_{j-1}})(\boldsymbol{D_{j-1}}) \subseteq \boldsymbol{D_j} \cap \boldsymbol{P_j^{-1}}(\{\boldsymbol{1}\}) \tag{3}$$

Here, $\boldsymbol{n}$ is the number of operations, and $\boldsymbol{j}$ indexes operation $\boldsymbol{f_j}$. The set $\boldsymbol{P_j^{-1}}(\{\boldsymbol{1}\}) = \{\boldsymbol{S} \in \boldsymbol{D_j} | \boldsymbol{P_j}(\boldsymbol{S}, \boldsymbol{\theta_j}) = \boldsymbol{1}\}$ contains the states satisfying the precondition of $\boldsymbol{f_j}$. Complete notation definitions are provided in Supplementary Table S1.

### Bounded loop

For a loop body spanning operations $\boldsymbol{f_a}$ through $\boldsymbol{f_b}$, define

$$\boldsymbol{B} = \boldsymbol{f_b}(\boldsymbol{\theta_b}) \circ \cdots \circ \boldsymbol{f_a}(\boldsymbol{\theta_a}), \qquad \boldsymbol{F_{loop}} = \boldsymbol{B^\tau} \tag{4}$$

Here, $\boldsymbol{B^\tau}$ denotes the $\boldsymbol{\tau}$-fold composition of the complete loop body $\boldsymbol{B}$, with $\boldsymbol{B^0} = \boldsymbol{I}$, where $\boldsymbol{I}$ is the the identity mapping. Let $\boldsymbol{x^{(0)}} = \boldsymbol{x_{in}}$ denote the loop-entry state and $\boldsymbol{x^{(j+1)}} = \boldsymbol{B}(\boldsymbol{x^{(j)}})$. Thus, $\boldsymbol{x^{(j)}}$ is the state after $\boldsymbol{j}$ completed executions of the loop body. Define $\boldsymbol{g_j} \coloneqq g(x^{(j)})$, where $\boldsymbol{g_j} = \boldsymbol{1}$ indicates that another iteration is required and $\boldsymbol{g_j} = \boldsymbol{0}$ indicates normal exit. The iteration count is

$$\boldsymbol{\tau} = \boldsymbol{min}(\{\boldsymbol{j} \in \{\boldsymbol{0}, \dots, \boldsymbol{N_{max}}\} \mid \boldsymbol{g_j} = \boldsymbol{0}\} \cup \{\boldsymbol{N_{max}}\}) \tag{5}$$

Reaching the upper bound while the continuation predicate remains true produces a non-empty loop diagnostic:

$$(\boldsymbol{\tau} = \boldsymbol{N_{max}}) \wedge (\boldsymbol{g_\tau} = \boldsymbol{1}) \Longrightarrow \boldsymbol{d_{loop}} \neq \emptyset \tag{6}$$

### Conditional branch

For *m* branches with mutually exclusive and collectively exhaustive guards, let $\boldsymbol{q_i}(\boldsymbol{x})$ denote the binary guard associated with branch $\boldsymbol{i}$.

$$\boldsymbol{q_i}: \boldsymbol{D_{cond}} \to \{\boldsymbol{0}, \boldsymbol{1}\} \quad (\boldsymbol{i} = \boldsymbol{1}, \dots, \boldsymbol{m}), \qquad \forall \boldsymbol{x} \in \boldsymbol{D_{cond}}: \sum_{i=1}^{m} \boldsymbol{q_i}(\boldsymbol{x}) = \boldsymbol{1} \tag{7}$$

For each input $\boldsymbol{x}$, let $\boldsymbol{i}^*$ denote the unique active branch index satisfying $\boldsymbol{q_{i^*}}(\boldsymbol{x}) = \boldsymbol{1}$.

The corresponding conditional operator is defined as:

$$\boldsymbol{F_{cond}}(\boldsymbol{x}) = \boldsymbol{f_{i^*}}(\boldsymbol{\theta_{i^*}})(\boldsymbol{x}) \tag{8}$$

**Parallel fork–join**

For a global state $\boldsymbol{x}$ and $p \geq 2$ parallel branches, let $\boldsymbol{\pi_{Di}}$ project $\boldsymbol{x}$ onto the branch-local objects and resources required by branch $\boldsymbol{i}$ ,and let $\boldsymbol{J}$ merge the resulting branch outputs at the synchronization point:

$$\boldsymbol{F_{par}}(\boldsymbol{x}) = \boldsymbol{J}\left(\boldsymbol{f_1}(\boldsymbol{\theta_1})\left(\boldsymbol{\pi_{D1}}(\boldsymbol{x})\right), \dots, \boldsymbol{f_p}(\boldsymbol{\theta_p})\left(\boldsymbol{\pi_{Dp}}(\boldsymbol{x})\right)\right), p \geq 2 \tag{9}$$

Let $\boldsymbol{W_i}$ and $\boldsymbol{L_i}$ denote the write and read sets of branch $\boldsymbol{i}$, respectively. Cross-branch write–read and write–write conflicts are excluded by

$$\forall \boldsymbol{i}, \boldsymbol{j} \in \{\boldsymbol{1}, \dots, \boldsymbol{p}\}, \boldsymbol{i} \neq \boldsymbol{j}: \ \boldsymbol{W_i} \cap \left(\boldsymbol{L_j} \cup \boldsymbol{W_j}\right) = \emptyset. \tag{10}$$

If two branch-local projections overlap, their outputs must agree on the shared state components before they are joined:

$$\begin{gathered} \forall \boldsymbol{i}, \boldsymbol{j} \in \{\boldsymbol{1}, \dots, \boldsymbol{p}\}, \boldsymbol{i} \neq \boldsymbol{j}, \boldsymbol{D_i} \cap \boldsymbol{D_j} \neq \emptyset: \\ \boldsymbol{\pi_{D_i \cap D_j}}(\boldsymbol{f_i}(\boldsymbol{\theta_i})\left(\boldsymbol{\pi_{Di}}(\boldsymbol{x})\right)) = \boldsymbol{\pi_{D_i \cap D_j}}(\boldsymbol{f_j}(\boldsymbol{\theta_j})\left(\boldsymbol{\pi_{Dj}}(\boldsymbol{x})\right)) \end{gathered} \tag{11}$$

Equation (10) permits shared read-only state while excluding conflicting cross-branch updates, whereas Equation (11) ensures deterministic merging of overlapping branch-local outputs. This formulation supports parallel operations on shared read-only state or distinct subsamples, followed by synchronization at the join point.

**Protected atomic interval**

For a protected interval *A* spanning operations $\boldsymbol{f_a}$ through $\boldsymbol{f_b}$, where $1 \leqslant a \leqslant b \leqslant n$, define

$$\boldsymbol{A} = \{\boldsymbol{f_a}, \dots, \boldsymbol{f_b}\}, \qquad \boldsymbol{F_A} = \boldsymbol{f_b}(\boldsymbol{\theta_b}) \circ \cdots \circ \boldsymbol{f_a}(\boldsymbol{\theta_a}) \tag{12}$$

Let $\boldsymbol{\sigma}(\boldsymbol{f_i})$ denote the unique position of operation instance $\boldsymbol{f_i}$ in the global execution schedule. The operations in $\boldsymbol{A}$ occupy consecutive schedule positions:

$$\boldsymbol{\sigma}(\boldsymbol{f}_{a+k}) = \boldsymbol{\sigma}(\boldsymbol{f}_{a}) + \boldsymbol{k}, \qquad \boldsymbol{k} = \mathbf{0}, \dots, \boldsymbol{b} - \boldsymbol{a} \tag{13}$$

No operation outside $\boldsymbol{A}$ may be scheduled within the interval bounded by its endpoints:

$$\forall \boldsymbol{h} \notin \boldsymbol{A}: \boldsymbol{\sigma}(\mathbf{h}) \notin [\boldsymbol{\sigma}(\boldsymbol{f}_{a}), \boldsymbol{\sigma}(f_{b})] \tag{14}$$

The figure label $\boldsymbol{F}_{lock}$ and the mathematical symbol $\boldsymbol{F}_{A}$ refer to the same protected, non-interleavable execution construct.

**Cross-references: Fig. 1c; Supplementary Tables S1 and S4; Supplementary Fig. S2; Supplementary Data 2.**

**Supplementary Note 4. Nested workflow algebra and serialization**

The expanded workflow in Fig. 1d composes the five operator classes within a single graph. Primitive functions form locally ordered sequences. A bounded outer loop controls repeated execution. The loop contains a fork–join region, within which one branch contains an inner loop, one contains a conditional branch and one contains a protected interval. The join synchronizes branch outputs before evaluation of the outer stopping predicate.

The executable representation preserves this hierarchy as an abstract syntax tree. Physical operations are namespace function calls. Parallel blocks are represented by ***parallel_execution*** containing named ***branch*** contexts. Protected intervals use ***atomic_lock***. Conditions use Python ***if/elif/else***, and bounded repetition uses ***while*** or ***for***. The converter recursively serializes these constructs into JSON without flattening their nesting.

The current converter produces a canonical structural string using SEQ, PAR, ATOMIC, IF, WHILE and FOR. Individual operations are indexed as $f_i$. The canonical string records topology rather than chemical threshold values. The JSON additionally records plan name, variables, ordered steps, workstation names, operation names, parameters, graph paths and switch mappings. Namespace calls retain workstation_eng_name, enabling deterministic routing back to a workstation-specific interface.

**Cross-references: Fig. 1d; Supplementary Fig. S2; Supplementary Tables S1 and S4; Supplementary Data 2.**

**Supplementary Note 5. Capability acquisition and Function-Skill admission**

The engineering pipeline converts laboratory capability descriptions into agent-readable and simulator-readable contracts. The supplied release contains 49 workstation specifications: 28 synthesis workstations, 5 reaction and testing workstations, 12 characterization workstations and 4 virtual control-flow workstations. The 45 physical workstations provide laboratory operations, whereas the 4 virtual modules provide parallel, conditional, loop and protected-interval constructs. Across these files, the current snapshot contains 63 declared Function-Skill functions. Shared operation names remain distinguishable through workstation namespaces.

Capability acquisition begins from a structured workstation-information file or manual authoring. The information schema includes workstation identity, operation descriptions, parameters, input and output constraints, state transitions, returns and file parameters. Natural-language Skill documents are admitted through a configurable 100-point assessment. The six required checks contribute 60 points, and a score of at least 60 is required for admission. Nine optional checks contribute the remaining 40 points. Virtual modules and workstations without material state receive documented relaxations for inapplicable checks.

Admitted Skills are converted into Function-Skill interface documents, ***_common.py*** and ***simulator.py***. Each interface function contains a typed signature and an eight-part docstring: Summary, Preconditions, Cautions, Args, Constraints, State Changes, Returns and Example. The converter centralizes shared constants, data classes, result classes, exception categories and namespace classes. The simulator implements the declared checks and state changes.

Fig. 2a–c separates development from use. Stages I–III construct, assess and convert the capability layer. Stage IV loads the control Skill, generates a recipe, analyzes and simulates the recipe, reviews it and returns diagnostics for regeneration when necessary. A generated workflow becomes a candidate for downstream dispatch only after the configured checks and workflow review pass.

The supplied release materials include Additional File 1 (source-code archive) and Additional File 2 (Function-Skill library), with 49 Skill documents, the shared runtime, simulator and conversion/assessment sources. No version tag or commit identifier is provided. Local archive checksums are **AF-01**: 59C0F68A82C3E7A9EF168F305658D62B5D0D70D2F7B12BB1DF97AAA1EFD430E7 and **AF-02**:

A3D71B6EB741F95900E3E5505E232B6BF9B1C80C5B5796D9C5F6ABB11DC855E
E.

**Cross-references: Fig. 2a,b; Supplementary Tables S5 and S6; Supplementary Figs. S3 and S4; Supplementary Data 1.**

**Supplementary Note 6. Static analysis and recipe optimization**

The recipe optimizer parses a generated Python recipe into an abstract syntax tree before simulation. Its security gate rejects syntax and constructs outside the bounded recipe language. The supplied implementation restricts imports to ***_common***, blocks dynamic execution and system, network, reflection and file-access functions, and validates function calls against names obtained from the Function-Skill library and common runtime.

The optimizer implements five cumulative levels. O0 performs the security gate. O1 adds recipe admission, normalization and flow analysis. O2 applies conservative AST simplification while preserving workstation-call order and experimental semantics. O3 can disable unused conditional outputs using laboratory-specific rules. O4 can merge compatible consecutive calls to the same workstation operation. Each transformed tree is checked again after modification.

The configuration supplied with the source uses O0 as the default. O0, O1 and O2 are enabled, whereas O3 and O4 are disabled. All four archived case-validation reports record O2 structural analysis as passed; no O3 or O4 transformation was applied.

The converter subsequently serializes the accepted AST into canonical JSON. Static acceptance establishes conformance to the encoded recipe language and workstation call inventory. It does not replace stateful checks, which are performed by the simulator, or scientific review, which evaluates whether the workflow addresses the stated intent.

**Cross-references: Fig. 2c; Supplementary Table S7; Supplementary Fig. S5; Supplementary Data 4; Additional File 1.**

**Supplementary Note 7. Stateful simulation and diagnostic-guided regeneration**

The simulator evaluates each operation against its executable state. For operation k, Fig. 2c defines the validation indicator

$V_k(SimState_t, \theta_k) = \Lambda_j \in M_k c_{kj}(SimState_t, \theta_k),$

where $M_k$ is the set of MUST constraints and $c_{kj}$ is constraint *j*. When all MUST constraints pass,

$V_k = 1: (SimState_{t+1}, \Delta_k) = (T_k(SimState_t, \theta_k), \emptyset),$

where $T_k$ is the encoded transition. When at least one MUST constraint fails,

$V_k = 0: (SimState_{t+1}, \Delta_k) = (SimState_t, \{e_{kj} \mid c_{kj} = 0\}).$

This validate-then-commit rule prevents a failed call from partially modifying the simulation state. The simulator continues checking compatible downstream information and returns a batch of diagnostics. SHOULD failures are retained separately as warnings.

Each diagnostic is an ErrorEntry containing function name, step number, unique code, category, message, expected value, actual value, remediation, parameter snapshot and MUST/SHOULD status. The supplied categories cover volume, container, status, parameter, temperature, time, speed, pressure, file, material, sample-state, workflow and general simulation failures. Supplementary Table S8 lists the reporting schema and categories.

The runner dynamically injects simulator implementations into the workstation namespace classes declared in ***_common.py***. It resets ***SimState***, executes ***generate_recipe()***, reports errors and warnings separately and returns acceptance only when the MUST-error list is empty. The virtual control-flow contexts record entry and exit events for protected and parallel blocks. The JSON converter preserves the nested topology for downstream adapters.

Fig. 2c is a schematic of the validate–diagnose–regenerate pattern rather than a case-specific archived regeneration trace.

**Cross-references: Fig. 2c; Supplementary Table S8; Supplementary Figs. S5 and S6; Supplementary Data 4; Additional File 1.**

**Supplementary Note 8. Fe coordination environment and alkaline OER workflow**

Extended Data Fig. 1a specifies a workflow to compare Fe-free $Ni(OH)_2$, co-precipitated Fe, post-impregnated Fe, citrate-complexed Fe and EDTA-complexed Fe while holding nominal Fe loading and electrochemical conditions constant. The recipe covers matched precursor addition, room-temperature mixing, staged drying and redispersion, XRD/infrared characterization and alkaline OER testing.

The attached case archive contains a 27-step linear workflow using 11 unique workstations. Its canonical representation is $SEQ(f_1(\theta_1),\ldots,f_{27}(\theta_{27}))$; no parallel branch or bounded loop is encoded because the same vial class is reused across the formulation routes. O2 structural analysis passed, simulation passed with 27 steps, zero MUST errors and zero SHOULD warnings, and workflow generation returned template 2701712185720832.

The case manifest specifies three independent batches, matched synthesis/OER aliquots and external confirmation of local Fe coordination by Fe K-edge XANES/EXAFS, XPS, ICP-OES and optional $^{57}Fe$ Mossbauer spectroscopy. The automated workflow therefore establishes capability-relative structure and simulation admissibility; it does not by itself establish the coordination assignment or measured OER performance.

The case archive supplies the experiment-design manifest, validated JSON plan, executable recipe, validation report and workflow-service response. The reported evidence is limited to automated semantic review, structural checks and stateful simulation; physical execution data are not included.

**Cross-references: Extended Data Fig. 1a; Supplementary Tables S9–S11; Supplementary Fig. S7; Supplementary Data 3; Additional File 3.**

**Supplementary Note 9. LDH–HMF synthesis, electrochemistry and multimodal analysis workflow**

Extended Data Fig. 1b specifies integrated synthesis and electrochemical testing of layered double hydroxide catalysts for alkaline 5-hydroxymethylfurfural oxidation. Catalyst preparation and electrode fabrication form a protected sequential interval. Electrochemical testing then produces solid-electrode and liquid-product objects that are routed to parallel contact-angle, XRD and liquid-phase product analyses. Their outputs are joined for integrated assessment.

The archived topology is $\boldsymbol{F_{hybrid}} = \boldsymbol{F_{par}} \circ \boldsymbol{F_{lock}} \circ \boldsymbol{F_{seq}}$. It contains a nine-operation protected preparation interval, sequential electrochemical evaluation, three object-specific analytical branches and an explicit post-branch join.

The attached case archive supplies the 22-step LDH-HMF JSON plan and Python recipe, with a nine-operation protected preparation interval and three parallel post-test branches. O2 analysis passed; simulation executed 31 steps with zero MUST errors and zero SHOULD warnings; workflow generation returned template 2695905526579200. The case-specific run sheet fixes 1.0 M KOH plus 10 mM HMF, while the validation report records adapter and sample-handoff assumptions.

The archive records NiFe-LDH preparation, electrode fabrication, CV/EIS/constant-potential testing, contact-angle, XRD and liquid-chromatography branches, followed by integrated evaluation. The reported evidence is limited to automated semantic review and stateful simulation; physical outcome data are not included. The contact-angle holder adapter, electrode scraping/transfer and electrolyte composition remain explicit execution assumptions.

**Cross-references: Extended Data Fig. 1b; Supplementary Tables S9–S11; Supplementary Fig. S8; Supplementary Data 3; Additional File 3.**

**Supplementary Note 10. Furfural hydrogenation and closed-loop GC optimization workflow**

Extended Data Fig. 1c specifies catalyst synthesis followed by electrocatalytic hydrogenation of furfural to furfuryl alcohol and closed-loop optimization of gas-chromatographic separation. The analytical loop adjusts method variables until the predefined peak-resolution criterion is met. The selected method is then applied to quantify furfural, furfuryl alcohol and major liquid by-products and to calculate conversion, yield and selectivity.

The archived topology is $\boldsymbol{F_{hybrid}} = \boldsymbol{F_{seq}} \circ \boldsymbol{F_{loop}}$. The full workflow must distinguish the chemistry branch from the method-optimization loop and define the join that makes the optimized method available to the quantification step. The loop requires an explicit objective, candidate parameter space, stopping predicate and maximum iteration count. Reaching the bound without satisfying the resolution predicate must return $\boldsymbol{d_{loop}}$, not a successful method.

The attached case archive supplies the exact furfural workflow specification, Python recipe, canonical JSON, GC method and batch files, and validation report. The workflow contains 20 physical nodes, an O2-passed bounded WHILE loop with $N_{max}$ = 6, pass/loop-back/fail-at-cap routes and a final quantification stage. Simulation passed with 23 executed steps, zero MUST errors and zero SHOULD warnings; workflow generation returned template 2696024588517376.

The closed-loop decision requires Rs ≥ 1.5 for both peak pairs, tailing ≤ 2.0 and retention-time RSD ≤ 1.0%. The validation report identifies the GC controller, measured metrics and method persistence as assumptions. No chromatograms or physical conversion/yield/selectivity outcomes are supplied, so the case supports workflow construction rather than measured chemical performance.

**Cross-references: Extended Data Fig. 1c; Supplementary Tables S9–S11; Supplementary Fig. S9; Supplementary Data 3; Additional File 3.**

**Supplementary Note 11. ZnO synthesis and adaptive spectroscopic characterization workflow**

Extended Data Fig. 1d specifies automated ZnO nanoparticle synthesis, centrifugal purification, drying, redispersion and parallel FTIR and UV–vis characterization. The UV–vis branch includes dilution and remeasurement until absorbance lies within a predefined quantitative interval. The FTIR and accepted UV–vis outputs are then joined.

The attached case archive supplies the ZnO Python recipe, canonical JSON, submitted workflow request, solid-weighing workbook and validation report. The workflow contains 27 physical nodes across 15 workstations, two parallel characterization branches and a bounded UV–vis dilution/remeasurement loop with $N_{max}$ = 3 and a 0.20–1.00 AU acceptance interval. O2 analysis passed; simulation executed 30 steps with zero MUST errors and zero SHOULD warnings; workflow generation returned template 2690180845338624.

The FTIR and UV–vis branches use disjoint container classes. The UV–vis guard applies the measured absorbance interval conceptually, while the supplied simulation uses a deterministic starting absorbance and a 0.5 dilution response.

**Cross-references: Extended Data Fig. 1d; Supplementary Tables S9–S11; Supplementary Fig. S10; Supplementary Data 3; Additional File 3.**

**Supplementary Note 12. Reproducibility and release package**

Reproduction requires a frozen software release, a frozen Function-Skill library and the exact artifacts for every reported workflow. The release should include the recipe optimizer, simulator runner, simulator implementation, common runtime, recipe converter, Function-Skill converter, quality assessor, configuration files, dependency specification and documentation. It should also include the four prompts, archived recipes, AST/JSON outputs, automated semantic-review reports and simulation logs.

The supplied package now includes Additional File 1 (source-code archive), Additional File 2 (49-workstation Function-Skill library), Additional File 3 (the four-case workflow archive) and Additional File 4 (release-page text). The local source archive contains the optimizer, simulator, converter, runtime and configuration files. A commit identifier, dependency lockfile and final non-sensitive model record remain unspecified.

The local release artifacts have been assigned archive-level SHA-256 values in Supplementary Data 4. Before public or reviewer release, the source archive must exclude credential-bearing configuration, internal object-storage URLs, temporary files and local paths. The AF-04 address is not treated as a persistent public link.

The case dialogues record the scientific prompts and generation dates in the four AF-03 folders. Exact provider/model, decoding parameters, random seeds and a deterministic reproduction statement are not present in the supplied package; those fields remain release metadata rather than inferred values.

Additional Files 1–3 are present locally. Additional File 4 contains an author-supplied internal release-page record, not a verified persistent public or reviewer-access repository. The three local archives are catalogued in Supplementary Table S12 and their checksums recorded in Supplementary Data 4. The supplied package does not include a dependency lockfile or final model metadata, and the source archive requires credential and internal-path sanitization before release.

Cross-references: Supplementary Tables S10–S12; Supplementary Data 4; Additional Files 1–4.

## Supplementary Tables

Supplementary Table S1. Authoritative notation used in Figs. 1 and 2, Extended Data Fig. 1 and the mathematical-definition document.

| Symbol | Definition | Scope or condition |
|---|---|---|
| $S_t$ | Typed research-object state at workflow step $t$ | Contains identity, sample, container, evidence and extensible object-level fields |
| $x$ | Generic state argument | Generic state argument supplied to loop, conditional and parallel operators |
| $t$ | Workflow-step index | Used in St and SimStatet |
| $f_i$ | Primitive operation $i$ | Workstation-bound state transition |
| i,j,k | Locally quantified integer indices | Index operations, branches, loop checks, constraints or schedule offsets as specified in each equation |
| $\theta_i$ | Parameters of operation $i$ | Typed and constrained by the corresponding Function Skill |
| $e_k$ | Non-negative error or convergence metric evaluated after iteration $k$ of a loop. | Used in $F_{loop}$; convergence is reached when $\boldsymbol{e_k \le \varepsilon}$. |
| $\varepsilon$ | Prespecified non-negative convergence tolerance; the loop terminates when $\boldsymbol{e_k \le \varepsilon}$. | Convergence threshold for $F_{loop}$; the loop continues while $\boldsymbol{e_k > \varepsilon}$. |
| $z$ | Observed measurement, diagnostic quantity, or state-derived value supplied to a conditional decision function. | Input quantity supplied to $Q(z)$ for evaluation of an $F_{cond}$ branch. |
| $Q(z)$ | Scalar decision metric evaluated from $\boldsymbol{z}$ for conditional branch selection. | Used in $F_{cond}$ ; the positive branch is selected when $\boldsymbol{Q(z) \ge Q_{thre}}$. |
| $Q_{thre}$ | Prespecified decision threshold; the positive branch is selected when | Decision threshold for ; separates the positive and negative branches. |
| $r$ | Current repetition count of the enclosing workflow. | Repetition counter for the enclosing workflow; compared with $\boldsymbol{R_{max}}$ after each repetition. |
| $R_{max}$ | Maximum permitted number of repetitions of the enclosing workflow before forced termination. | Hard repetition bound; execution terminates when $\boldsymbol{r \ge R_{max}}$ , otherwise the enclosing workflow repeats. |
| $P_i$ | Precondition predicate for operation $i$ | Operation is admissible when $P_i(S_t,\theta_i)=1$ |

| Symbol | Definition | Scope or condition |
|---|---|---|
| $\boldsymbol{P_j^{-1}}(\{\mathbf{1}\})$ | Set of states satisfying precondition $P_j$ | $\boldsymbol{P_j^{-1}}(\{\mathbf{1}\}) = \{\boldsymbol{S} \in \boldsymbol{D_j} \mid \boldsymbol{P_j}(\boldsymbol{S}, \boldsymbol{\theta_j}) = \mathbf{1}\}$ |
| $D_i$ | Domain accepted by operation or branch *i* | Used as an accepted input domain in Eq. (3) and as a branch-local projection target in Eqs. (9) and (11) |
| $F_{seq}$ | Sequential composition | Defined in Eq. (2); adjacent-domain compatibility is specified by Eq. (3) |
| $n$ | Number of operations in the considered ordered sequence | Used in Eqs. (2) and (3) and in $1 \leq a \leq b \leq n$ |
| $B$ | Transformation corresponding to one complete loop-body execution | Eq. (4) |
| $I$ | Identity state transformation | $I(x)=x$ |
| $x_{in}$ | Loop-entry state | $x^{(0)}=x_{in}$ |
| $x^{(j)}$ | State after j completed loop-body executions | $\boldsymbol{x}^{(j+1)} = \boldsymbol{B}(\boldsymbol{x}^{(j)})$ |
| $F_{loop}$ | Bounded loop composition | Eqs. (4)–(6) |
| $g_j$ | Loop continuation predicate after check *j* | One continues; zero exits |
| $\tau$ | Realized loop count | Bounded by $N_{max}$ |
| $N_{max}$ | Maximum permitted loop count | Produces $d_{loop}$ if continuation remains true |
| $d_{loop}$ | Non-convergence diagnostic | Non-empty under Eq. (6) |
| $q_i(x)$ | Binary guard associated with conditional branch i | exactly one guard is active for each admissible input |
| $i^*$ | Active conditional-branch index for the current input x | Unique index satisfying $\boldsymbol{q}_{\boldsymbol{i}^*}(\boldsymbol{x}) = \mathbf{1}$ |
| $F_{cond}$ | Conditional composition | Unique active branch under Eqs. (7) and (8) |
| $m$ | Number of conditional branches | Exactly one $\boldsymbol{q_i(x)}$ equals one |
| $F_{par}$ | Parallel fork–join composition | Eqs. (9)–(11) |

| Symbol | Definition | Scope or condition |
|---|---|---|
| $p$ | Number of parallel branches | $p \geq 2$ |
| $\pi_{Di}$ | Projection onto branch $i$'s required state and resources | Applied before branch execution |
| $J$ | Join operator | Merges consistent branch outputs |
| $W_i$ | Write set of branch $i$ | Used in Eq. (10) |
| $L_i$ | Read set of branch $i$ | Used in Eq. (10); retain this symbol from the supplied definition |
| $A$ | Protected atomic interval $\{f_a,\ldots,f_b\}$ | Eq. (12) |
| $F_A$ | Composition inside protected interval $A$ | Same control-flow class as figure label $F_{lock}$ |
| $F_{lock}$ | Figure label for protected non-interleavable composition | Use consistently with $F_A$ |
| $\sigma(f)$ | Position of operation $f$ in the schedule | Eqs. (13) and (14) |
| $h$ | Scheduled operation instance outside protected interval A | Used in Eq. (14) |
| $F_{hybrid}$ | Workflow composed from two or more operator classes | Used in Fig. 1d and Extended Data Fig. 1 |
| $k$ | Wash-loop iteration index | k=1,…,τ; used in $\boldsymbol{S_i^{(k)}}$ *for states generated during the* **k***-th execution of the wash-loop body* |
| $S_i^{(k)}$ | Research-object state at workflow node $i$ during wash-loop iteration $k$ | Applied to $i \in \{4,5,6,7,8,9\}$ |
| $SimState_t$ | Executable simulator state at step $t$ | Bounded executable projection of the conceptual research-object collection $S_t$ |
| $V_k$ | MUST-validation indicator for operation $k$ | Conjunction of $c_{kj}$ |
| $M_k$ | Set of MUST constraints for operation $k$ | Used by $V_k$ |
| $c_{kj}$ | Constraint $j$ for operation $k$ | Boolean predicate |
| $T_k$ | Encoded state transition for operation $k$ | Committed only when $V_k=1$ |

| Symbol | Definition | Scope or condition |
| --- | --- | --- |
| $\Delta_k$ | Structured diagnostic set returned for operation $k$ | Empty on success; contains failed constraints on failure |
| $e_{kj}$ | Diagnostic associated with failed $c_{kj}$ | Included when $c_{kj}=0$ |

Supplementary Table S2. Conceptual research-object fields and their implemented simulation projection.

| **Component** | **Conceptual fields in Fig. 1b** | **Current executable fields** | **Evidence or limitation** |
|---|---|---|---|
| Identity | Object identifier and provenance links | Container key combines type and identifier; step log retains operation order | Object-level provenance beyond container identity requires case JSON |
| Sample | Identity, composition, phase, mass, volume and extensible properties | sample_status, sample_volume_ml, sample_mass_g | Composition is represented in recipe parameters or case records, not as a general SimState field |
| Container | Type, shape, capacity, lid or holder state and location | container_id, container_type, container_status, location | Geometry and capacity limits are encoded in Skill constraints and shared constants |
| Evidence | Temperature, pH, colour, XRD, UV–vis and other outputs | Typed result objects, test_result_file, logs and diagnostics | Operation-specific evidence schemas are recorded in typed result classes and case JSON records |
| Capability binding | Accepted container, phase or volume, parameter limits and expected effects | Function namespace, typed signature, preconditions, MUST/SHOULD constraints and state changes | Defined in Function-Skill documents |
| State history | $S_t \rightarrow S_{t+1}$ with retained provenance | SimState.logs, step counter and committed container state | Stepwise history is retained in SimState.logs and the case JSON records |

Supplementary Table S3. Function-Skill contract schema.

| Contract element | Required content | Executable realization | Acceptance requirement |
|---|---|---|---|
| Workstation | Unique English namespace, Chinese name, module and identifier | Namespace class in _common.py | Namespace must resolve to a registered simulator function |
| Summary | One-sentence operation purpose | Docstring | Must agree with source capability description |
| Preconditions | Accepted object, container and sample states | require_* checks and operation-specific checks | Every declared MUST precondition must be implemented |
| Cautions | Cross-step effects and device-specific restrictions | Documentation and automated checks | Must not be converted into hidden simulator behavior |
| Args | Typed values with unit-bearing names and enumerations | Function signature and data classes | Values must parse and fall within documented types |
| Constraints | MUST/SHOULD Boolean conditions with failure categories | record_error and record_warning | All feasible checks are collected in one pass |
| State Changes | Explicit target fields and effects | Validate-then-commit update | No state update is allowed after a failed MUST rule |
| Returns | StepResult or operation-specific subclass | Typed result object | Data-producing operations include evidence references |
| Example | Valid, internally consistent call | Skill document | Must use a registered namespace and legal values |
| Raises or diagnostics | Category, message, expected, actual and remediation | ErrorEntry | MUST and SHOULD records remain separate |

Supplementary Table S4. Composition operators, invariants and code representation.

| **Operator** | **Formal definition** | **Required invariant** | **Python representation** | **Canonical JSON token** |
|---|---|---|---|---|
| Sequence | Eqs. (2), (3) | Each output lies in the next domain and each precondition holds | Ordered statements | SEQ(...) |
| Bounded loop | Eqs. (4)–(6) | Explicit continuation predicate, finite $N_{max}$, diagnostic at non-convergence | while or for | WHILE(...) or FOR(...) |
| Conditional | Eqs. (7), (8) | Exactly one branch is selected | if/elif/else | IF(...) |
| Parallel fork–join | Eqs. (9)–(11) | No write conflict and agreement on shared state | parallel_execution with branch contexts | PAR(...) |
| Protected interval | Eqs. (12)–(14) | Contiguous scheduled positions with no external interleaving | atomic_lock context | ATOMIC(...) |

Supplementary Table S5. Capability inventory in the supplied Function-Skill snapshot.

| Module | Workstations, grouped by declared capability | Workstation count | Declared functions |
|---|---|---|---|
| Synthesis | Container storage; general material; liquid handling at 1 ml, 4 channels and 5 ml in multiple versions; liquid pouring; drying; single- and multi-channel solid dosing; solid transfer; centrifugation; cleaning and dispensing; cooling; heat-resistant container supply; heated and room-temperature stirring; muffle furnace; plate storage; spectroscopy holder preparation; ultrasonic dispersion and liquid handling; intelligent photocatalysis container transfer; purification | 28 | 41 |
| Reaction and testing | Dual-station electrochemistry; high-temperature/high-pressure microreaction; photocatalysis V1 and V2; post-reaction processing | 5 | 6 |
| Characterization | Dark-box imaging; fluorescence; gas chromatography; high-speed imaging; infrared; contact angle; illumination and membrane clamping; liquid chromatography; microplate reader; spectroscopy container transfer; UV–visible spectroscopy; XRD | 12 | 13 |
| Virtual control flow | Atomic lock; parallel branch; conditional jump; loop | 4 | 3 explicit context functions plus native Python control |
| **Total** | 45 physical and 4 virtual workstation specifications | **49** | 63 |

Supplementary Table S6. Natural-language Skill quality-assessment rubric.

| Check | Maximum score | Requirement |
|---|---|---|
| R1 Workstation name/title | 10 | Name, code and primary title |
| R2 Operation description | 10 | Explicit operation section and purpose |
| R3 Parameter information | 15 | Parameter names, types, levels, units and constraints |
| R4 Initial state | 10 | Container type, container state and sample state where applicable |
| R5 State update | 10 | Explicit effect or state-transition mapping |
| R6 Constraint rules | 5 | Business and parameter constraints |
| **Required subtotal** | **60** | Admission threshold is at least 60 |
| O1 Frontmatter completeness | 5 | Name, description, module and code |
| O2 Similar-workstation distinction | 5 | Version or capability distinction |
| O3 Overall workflow constraints | 5 | Required predecessor or process dependency |
| O4 Parameter-constraint format | 5 | Explicit ranges, enumeration or conditional dependency |
| O5 Nested-parameter expansion | 5 | Structured nested inputs |
| O6 File parameters | 3 | Template, format and transfer requirements |
| O7 Return values | 5 | Typed output schema |
| O8 State-transition completeness | 4 | Coverage of declared initial-state fields |
| O9 Example completeness | 3 | Valid example values and call |
| **Optional subtotal** | **40** | Used to assess conversion completeness |

Supplementary Table S7. Recipe optimizer levels and supplied configuration.

| Level | Implemented behavior | Supplied configuration | Final case-use evidence |
|---|---|---|---|
| O0 | Syntax and attack-surface gate; import, node, name and function-call restrictions | Enabled; default | Included in the O2 checks recorded for all four cases; no separate command line supplied |
| O1 | O0 plus generate_recipe() admission, normalization, flow warnings and merge suggestions | Enabled | Included in the O2 admission and flow checks |
| O2 | O1 plus conservative AST simplification; no workstation-call reordering | Enabled | Passed for Fe coordination/OER, LDH-HMF, furfural and ZnO |
| O3 | O2 plus disabling of unused conditional outputs under configured rules | Disabled | Disabled; no case evidence of use |
| O4 | O3 plus compatible consecutive-call merging | Disabled | Disabled; no case evidence of use; advisory merges were not applied |

Supplementary Table S8. Structured simulation diagnostics.

| Field or category | Meaning | Required reporting |
| --- | --- | --- |
| function_name | Operation that produced the diagnostic | Registered namespace operation |
| step_number | Position in simulated workflow | Integer step identifier |
| code | Unique machine-readable error code | Stable across reruns of the same release |
| category | Diagnostic class | One of the release categories below |
| message | Observed failure | Current value and expected condition |
| expected, actual | Normalized comparison | Explicit where applicable |
| remediation | Proposed repair | Parameter adjustment or missing/reordered operation |
| params_snapshot | Relevant call arguments | Sufficient to reproduce the failure |
| is_must | Blocking status | MUST enters errors; SHOULD enters warnings |
| Volume and volume-constraint categories | Insufficient or excessive sample or transfer volume | VolumeError, VolumeConstraintError |
| Container and material categories | Missing, incompatible or excessive container/material | ContainerConstraintError, MaterialConstraintError |
| State categories | Invalid container or sample state | StatusError, SampleStatusError |
| Parameter categories | Missing, out-of-range, dependent or otherwise invalid parameter | ParameterError, ParameterRangeError, ParameterConstraintError, ParameterDependencyError, ParameterMissingError |
| Physical-limit categories | Temperature, time, speed or pressure failure | TemperatureError, TimeError, SpeedError, PressureError |
| File categories | Invalid file or template input | FileFormatError, FileConstraintError |
| Workflow and simulation categories | Missing, unordered or otherwise invalid workflow element | WorkflowError, SimulationError |

Supplementary Table S9. Extended Data Fig. 1 scientific intents and displayed operator topologies.

| Panel | Scientific intent | Displayed algebra | Required evidence | Reconciliation status |
| --- | --- | --- | --- | --- |
| Extended Data Fig. 1a | Fe coordination environment modulation and alkaline OER | $\boldsymbol{F}_{hybrid} = \boldsymbol{F}_{seq}$ | Five Fe-introduction routes, matched OER aliquots, XRD/IR outputs and external coordination confirmation | 27-step linear archive; O2/simulation/workflow generation passed; coordination assignment and physical OER data remain external |
| Extended Data Fig. 1b | LDH synthesis, HMF oxidation and multimodal analysis | $\boldsymbol{F}_{hybrid} = \boldsymbol{F}_{par} \circ \boldsymbol{F}_{lock} \circ \boldsymbol{F}_{seq}$ | Protected endpoints, object projections, parallel branches and join | Exact archive supplied; O2 and simulation passed with zero errors; adapter assumptions and physical outcomes remain explicit |
| Extended Data Fig. 1c | Furfural hydrogenation and closed-loop GC optimization | $\boldsymbol{F}_{hybrid} = \boldsymbol{F}_{seq} \circ \boldsymbol{F}_{loop}$ | Objective, parameter space, stopping predicate, bound and method trace | Exact archive supplied; bounded loop and O2/simulation passed; chromatographic measurements remain assumptions |
| Extended Data Fig. 1d | ZnO synthesis and adaptive FTIR/UV–visible analysis | $\boldsymbol{F}_{hybrid} = \boldsymbol{F}_{par} \circ \boldsymbol{F}_{loop} \circ \boldsymbol{F}_{cond} \circ \boldsymbol{F}_{seq}$ | UV–vis condition, dilution action, repeat bound and join | Exact archive supplied; bounded ultraviolet–visible loop and nested conditional are explicit in canonical JSON |

Supplementary Table S10. Required per-case artifact and graph-metric record.

| Record | Fe coordination/OER | LDH–HMF | Furfural | ZnO |
|---|---|---|---|---|
| Exact prompt | Dialogue history.md; exact Fe coordination/OER prompt | Dialogue history.md; exact LDH-HMF prompt | Dialogue history.md; exact furfural/GC prompt | Dialogue history.md; exact ZnO prompt |
| First-pass recipe | One archived recipe; no separate v1/final pair | One archived recipe; no separate v1/final pair | One archived recipe; no separate v1/final pair | One archived recipe; no separate v1/final pair |
| Final recipe | Same archived recipe; no separate final pair | Same archived recipe; no separate final pair | Same archived recipe; no separate final pair | Same archived recipe; no separate final pair |
| Canonical AST/JSON | experiment_plan.json; SEQ($f_1$–$f_{27}$) | recipe_ldh_hmf_oxidation.json; ATOMIC + PAR | recipe_furfural_furfuryl_alcohol_gc_loop.json; WHILE | recipe_ZnO_nanoparticle_adaptive_spectroscopy.json; PAR + WHILE |
| Primitive-operation nodes | 27 | 22 | 20 | 27 |
| Unique physical workstations | 11 | 11 | 11 | 15 |
| Sequence blocks | 1 | 5 | 2 | 4 |

| Record | Fe coordination/OER | LDH–HMF | Furfural | ZnO |
| --- | --- | --- | --- | --- |
| Loop blocks and bounds | 0 | 0 | 1 ($N_{max}$ = 6) | 1 ($N_{max}$ = 3) |
| Conditional blocks | 0 | 1 | 1 | 2 |
| Parallel branches | 0 | 3 | 0 | 2 |
| Protected intervals | 0 | 1 ($f_1$–$f_9$) | 0 | 0 |
| MUST errors before regeneration | 0 | 0 | 0 | 0 |
| MUST errors after regeneration | 0 | 0 | 0 | 0 |
| SHOULD warnings after regeneration | 0 | 0 | 0 | 0 |

Supplementary Table S11. Automated semantic review and simulation outcomes.

| Case | Automated semantic review | Stateful simulation |
|---|---|---|
| Fe coordination/OER | Passed with warnings; external coordination confirmation remains outside the automated workflow. | Passed; 27 steps, 0 MUST errors, 0 SHOULD warnings. |
| LDH-HMF | Passed with warnings; adapter and sample-handoff assumptions are recorded. | Passed; 31 steps, 0 MUST errors, 0 SHOULD warnings. |
| Furfural-GC | Passed with assumptions; GC metric publication and method persistence are external controller assumptions. | Passed; 23 steps, 0 MUST errors, 0 SHOULD warnings. |
| ZnO spectroscopy | Passed with information; absorbance and dilution response are simulation controls. | Passed; 30 steps, 0 MUST errors, 0 SHOULD warnings. |

Supplementary Table S12. Software, model and release manifest.

| Artifact | Role | Supplied status |
|---|---|---|
| recipe_optimizer.py | Security, admission and AST optimization | AF-01 source archive; local archive hash recorded below |
| run_simulation.py | Runtime registration and simulation entry point | AF-01 source archive |
| references/simulator.py | Constraint checks and state transitions | AF-01 source archive |
| references/_common.py | Shared constants, types, results and namespaces | AF-01 source archive |
| recipe_converter.py | Python/JSON conversion and canonical form | AF-01 source archive |
| skills_to_functions.py | Natural-language to Function-Skill conversion | AF-01 source archive |
| skill_quality_assessor.py | Skill admission | AF-01 source archive |
| optimizer_config.json | Enabled optimizer levels | AF-01 source archive; O0–O2 enabled, O3–O4 disabled |
| o3_conditional_rules.yaml | Conditional-output optimization rules | AF-01 source archive |
| domain_validation_rules.yaml | Laboratory-specific validation rules | AF-01 source archive |
| quality_scoring_config.json | Skill admission rubric | AF-01 source archive; threshold 60 |
| Python and dependencies | Execution environment | No lockfile in supplied package |
| Generative model | Recipe or Skill generation | AF-03 dialogues and validation reports |
| Source-code release | Reproduction package | AF-01 Source-code archive.zip |
| Function-Skill release | Capability package | AF-02 Function-Skill library.zip |
| Extended Data Fig. 1 case archive | Exact prompts, recipes and logs | AF-03 Extended Data Fig. 1 workflow archive.zip |
| Release-page record | Internal provenance record; not a verified persistent public/reviewer route | AF-04 Release page.txt |

## Supplementary Figures

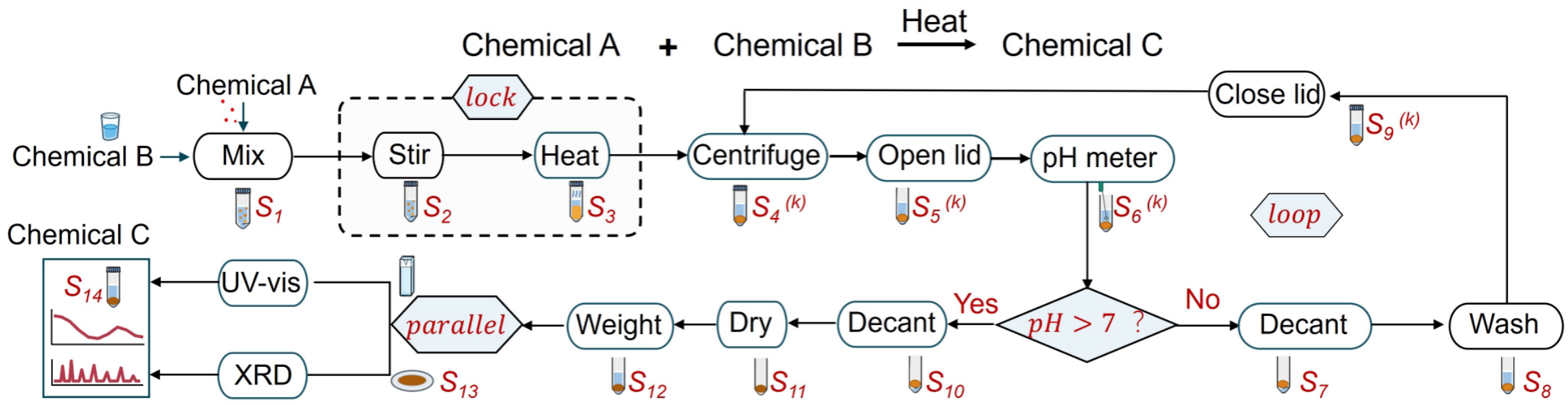

**Field-level state changes throughout the representative workflow**

| State | Triggering operation | Sample state | Mass/volume state | Container and lid state | Evidence appended / principal change |
|---|---|---|---|---|---|
| $S_1$ | $f_1$: mixing | Suspension of A and B | Total mass/volume | 50mL tube; lid closed | Mixing completed; sample identity and provenance initialized |
| $S_2$ | $f_2$: stirring | Homogenized suspension | Unchanged | 50mL tube; lid closed | Stirring time and speed appended |
| $S_3$ | $f_3$: heating | Heated reaction mixture | Unchanged | 50mL tube; lid closed | Heating temperature and duration appended |
| $S_4$ | $f_4$: centrifuge | Liquid-solid separated | Unchanged | 50mL tube; lid closed | Centrifugation speed and duration appended |
| $S_5$ | $f_5$: open lid | Liquid-solid mixture | Unchanged | 50mL tube; lid open | Lid-state transition appended |
| $S_6$ | $f_6$: pH measurement | Liquid–solid mixture | Unchanged | 50mL tube; lid open | Measured pH appended as evidence |
| $S_7$ | $f_7$: decant after pH≤7 | Retained solid | Supernatant removed | 50mL tube; lid open | Conditional result and decanted volume appended |
| $S_8$ | $f_8$: wash | Washed suspension | Wash liquid added | 50mL tube; lid open | Washing reagent, volume and cycle number appended |
| $S_9$ | $f_9$: close lid | Washed suspension prepared for repeated separation | Unchanged | 50mL tube; lid closed | Lid closure and loop-back event appended |
| $S_{10}$ | $f_{10}$: decant after pH>7 | Final washed solid retained; supernatant removed | Supernatant removed | 50mL tube; lid open | Passing decision and final decant record appended |
| $S_{11}$ | $f_{11}$: dry | Dry solid | Solvent removed | 50ml tube; lid open | Drying temperature and duration appended |
| $S_{12}$ | $f_{12}$: weigh | Dry solid aliquot | Measured mass | 50ml tube; lid open | Mass measurement appended as evidence |
| $S_{13}$ | $f_{13}$: parallel sample preparation/fork | Characterized solid | Unchanged | XRD holder and quartz cuvette | Branch identities, aliquot masses and parent–child provenance appended |
| $S_{14}$ | $f_{14}$: parallel characterization and join | Characterized Chemical C; material state nominally unchanged | Unchanged except for any consumed characterization aliquot | Characterization holders; branches completed and joined | XRD and UV–Vis results appended and combined into the final evidence record |

Supplementary Fig. S1 | Stateful synthesis–characterization workflow and field-level state changes. The workflow includes a protected sequence, a pH-dependent wash loop, and parallel UV-vis/XRD analysis. States S1-S14 are research-object states summarized in the table below. Superscript *k* denotes the wash-loop iteration index (k=1,…,τ); $S_i^{(k)}$ represents the research-object state at workflow node *i* during iteration *k*. States $S_4^{(k)}$ – $S_9^{(k)}$ belong to the repeated loop body, whereas $S_{10}$ is reached after the exit condition $pH(S_6^{(k)}>7)$ is satisfied.

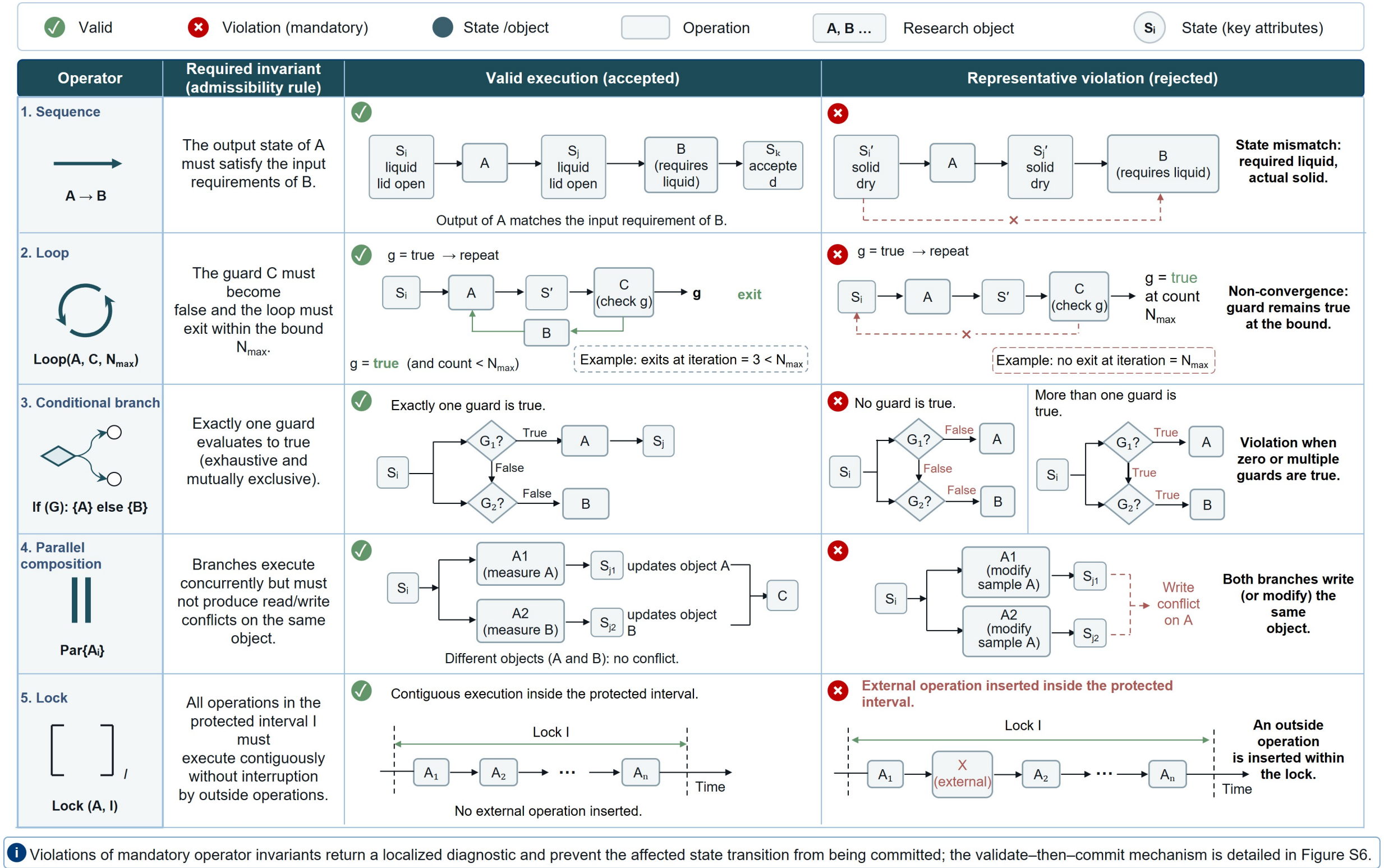

Supplementary Fig. S2 | Admissibility conditions and representative violations of the five workflow operators. Valid and rejected execution patterns are illustrated for sequence, loop, conditional, parallel, and protected-interval operators. Each operator is defined by a mandatory structural invariant; violations trigger a localized diagnostic and prevent the corresponding state transition from being committed.

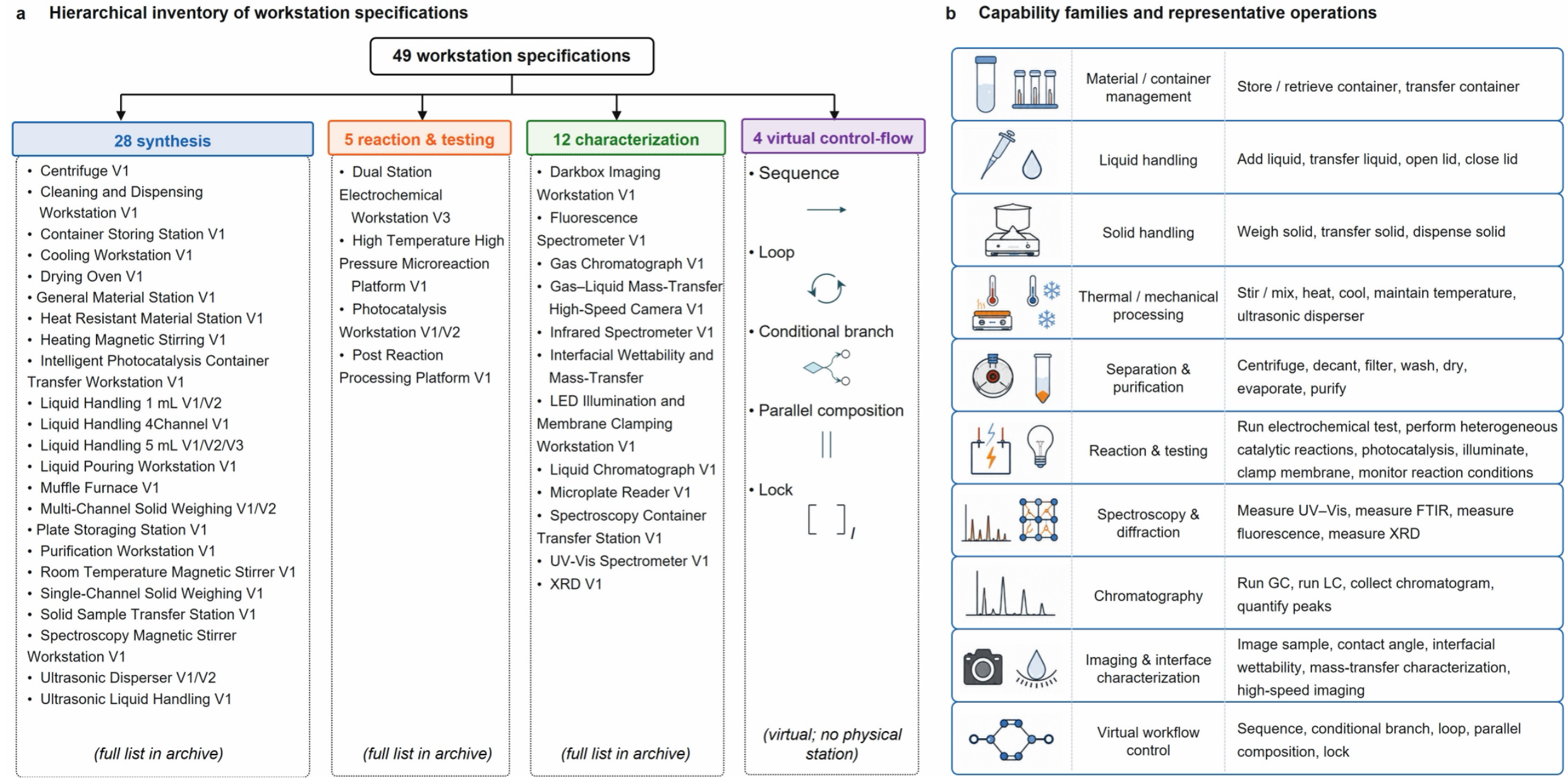

Supplementary Fig. S3 | Hierarchical inventory of workstation specifications and capability families. (a) The 49 workstation specifications are organized into synthesis, reaction and testing, characterization, and virtual control-flow categories, with representative workstations or operators shown. (b) The corresponding capability taxonomy summarizes representative operations spanning material and container management, sample handling, processing, separation, reaction, characterization, and workflow control.

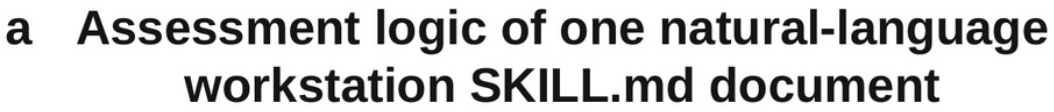

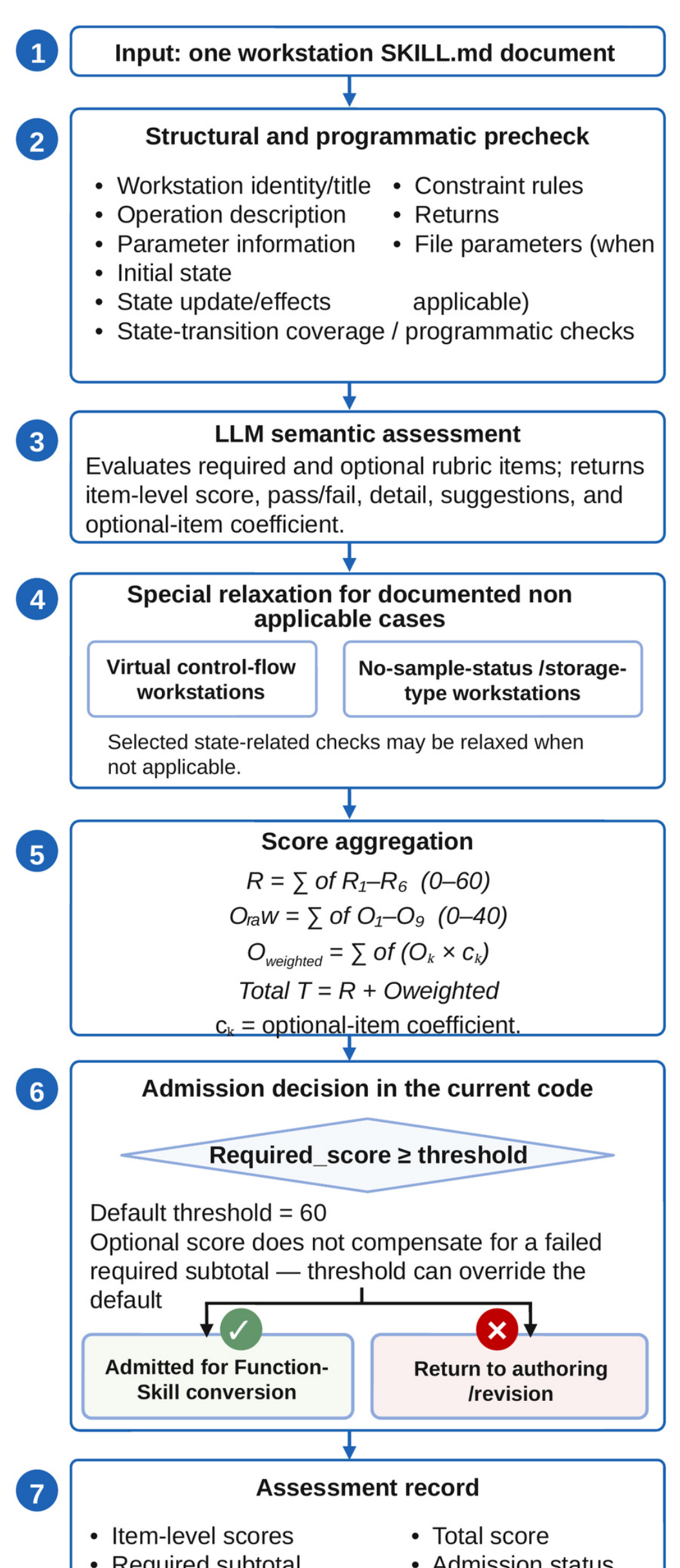

**b Rubric items defined in Supplementary Table S6**

| Required checks (60 points) | | |
|---|---|---|
| **R1** | Workstation name/title | 10 |
| **R2** | Operation description | 10 |
| **R3** | Parameter information | 15 |
| **R4** | Initial state | 10 |
| **R5** | State update | 10 |
| **R6** | Constraint rules | 5 |

| Optional checks (40 points) | | |
|---|---|---|
| **O1** | Frontmatter completeness | 5 |
| **O2** | Similar-workstation distinction | 5 |
| **O3** | Overall workflow constraints | 5 |
| **O4** | Parameter-constraint format | 5 |
| **O5** | Nested-parameter expansion | 5 |
| **O6** | File parameters | 3 |
| **O7** | Return values | 5 |
| **O8** | State-transition completeness | 4 |
| **O9** | Example completeness | 3 |

The quality score measures document completeness /conversion readiness; it is not an experimental performance metric.

Supplementary Fig. S4 | Natural-language Skill quality assessment and admission logic. (a) Workflow for structural checks, LLM-based semantic assessment, special-case handling, score aggregation, threshold-based admission, and assessment recording. (b) Required and optional rubric items used for Skill quality scoring, with a total score of 100 points.

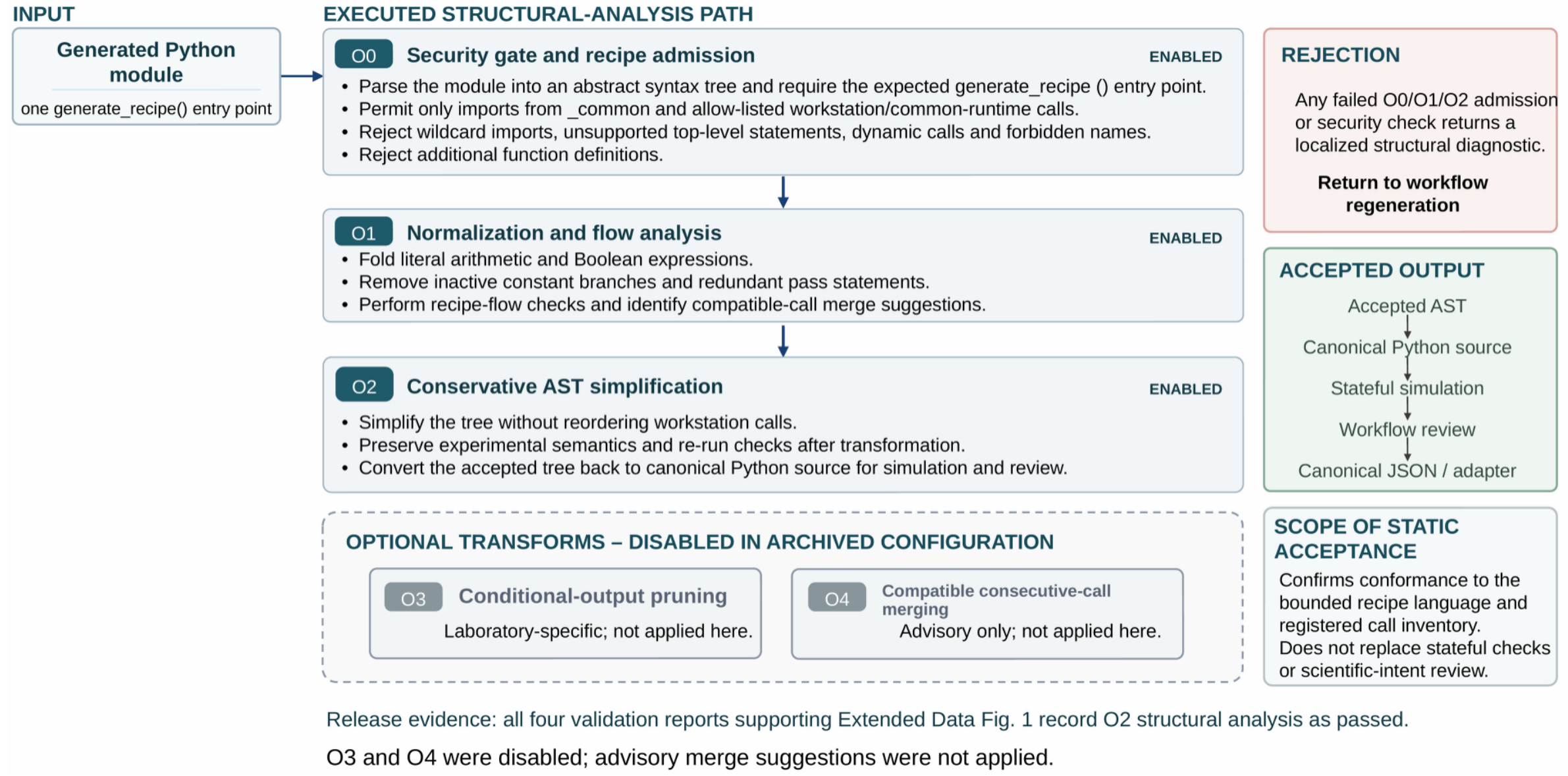

Supplementary Fig. S5 | Static analysis and conservative optimization of generated Python recipes. The workflow covers security admission, normalization, conservative AST simplification, and optional domain-aware optimization, with rejected recipes returned for regeneration and accepted recipes serialized for simulation and review. O0–O2 are enabled, whereas O3–O4 are disabled in the archived configuration. The workflow and configuration are aligned with Supplementary Note 6 and Supplementary Table S7.

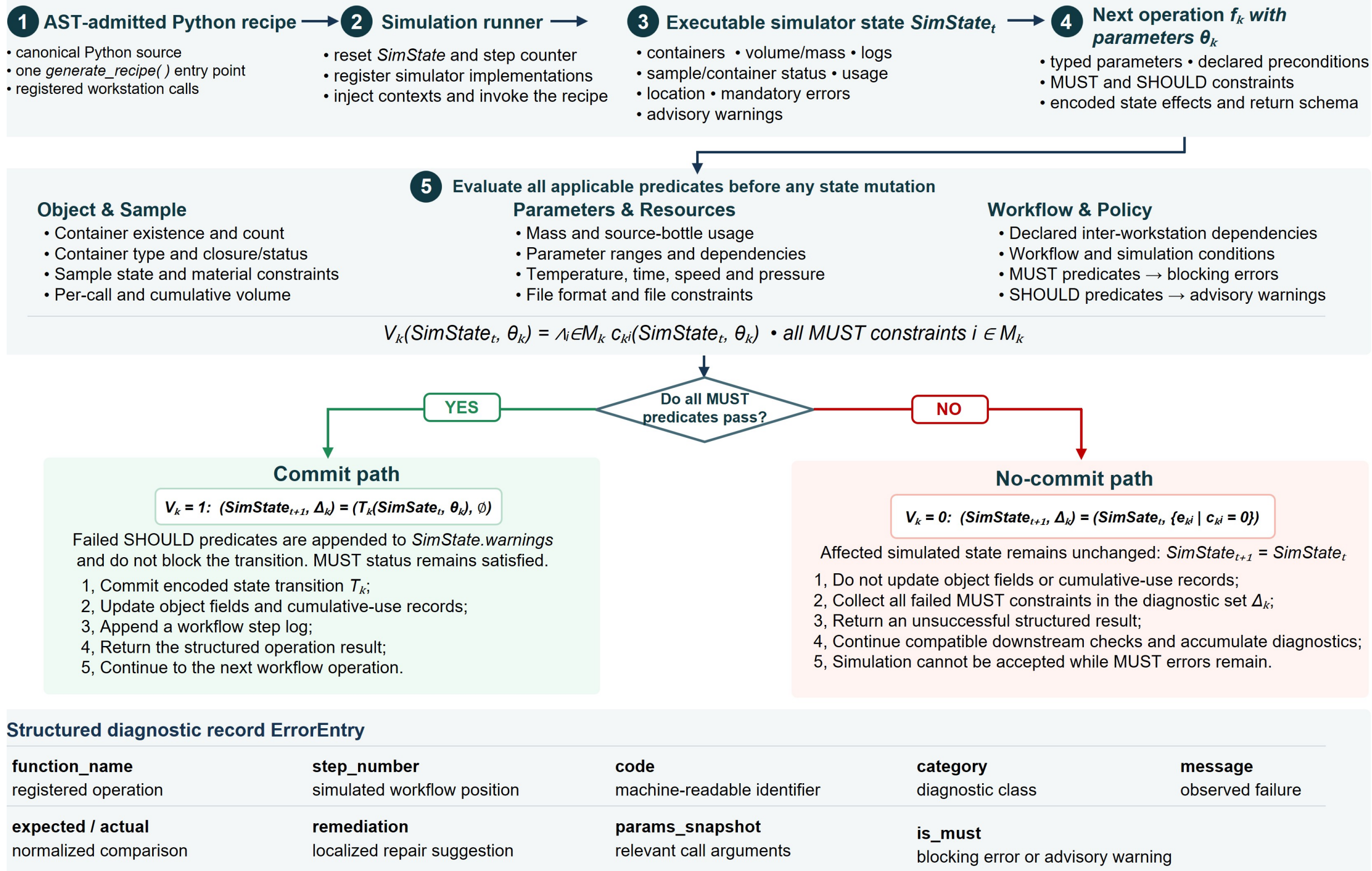

Supplementary Fig. S6 | Validate-then-commit stateful simulation and structured diagnostic reporting. Following static recipe admission, each operation is dynamically evaluated against all applicable MUST and SHOULD predicates before any state mutation. Operations satisfying all MUST constraints are committed, whereas failed MUST constraints trigger a no-commit path that preserves the current state and records structured diagnostics in ErrorEntry; SHOULD violations are retained as non-blocking warnings. Here, $SimState_t$ is the bounded executable projection of the global laboratory state $\mathcal{L}_t$. The schematic illustrates the operation-level validation logic and contains no case-specific values.

## Fe coordination environment and alkaline OER workflow

### a. Scientific intent and comparison design

**Objective:** Design an experiment to investigate how the existing form and coordination environment of Fe modulate the alkaline oxygen evolution reaction (OER) performance of NiFe-based catalysts. NiFe-based catalytic materials with different Fe introduction methods and distinct Fe coordination environments are required to be fabricated.

| Fe-free control | Co-precipitated Fe | Post-impregnated Fe | Citrate-mediated Fe | EDTA-mediated Fe |
|---|---|---|---|---|
| Fe-free reference | Lattice-associated Fe-O-Ni hypothesis | Surface-enriched FeOx/FeOOH-like hypothesis | Ligand-modulated hypothesis and incorporation | Stronger chelation-modulated Fe availability |

**External coordination confirmation:** Fe K-edge XANES/EXAFS • XRD • ICP-OES • optional $^{57}$Fe Mossbauer spectroscopy.

### b. Workstation architecture

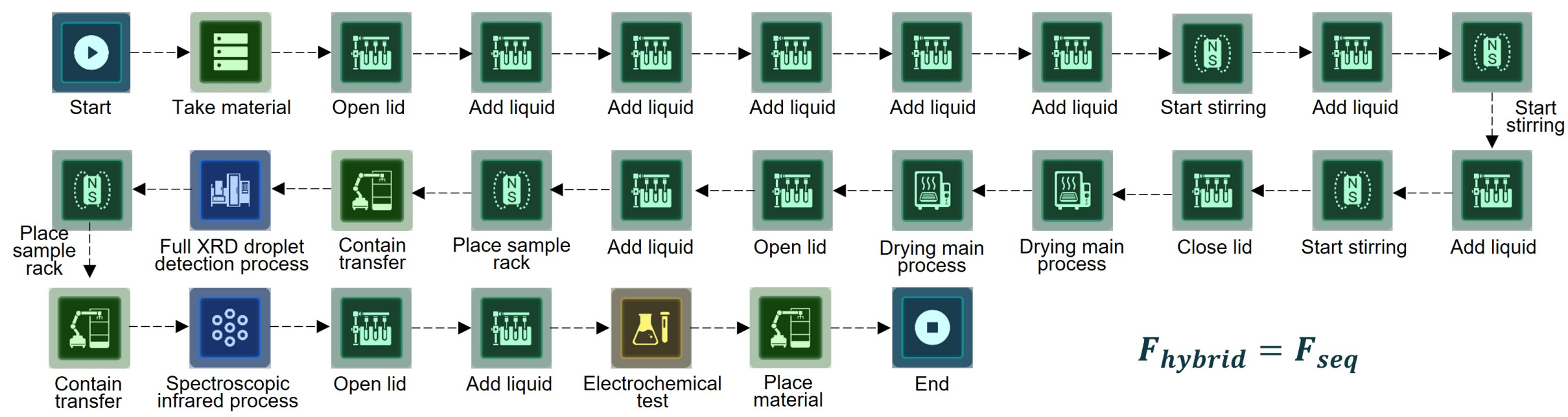

Five matched formulation routes are executed as a linear sequence because the same sample-vial container class is reused. Working coordination motifs remain hypotheses until confirmed by the external measurements listed above.

### c. Validation and archived record

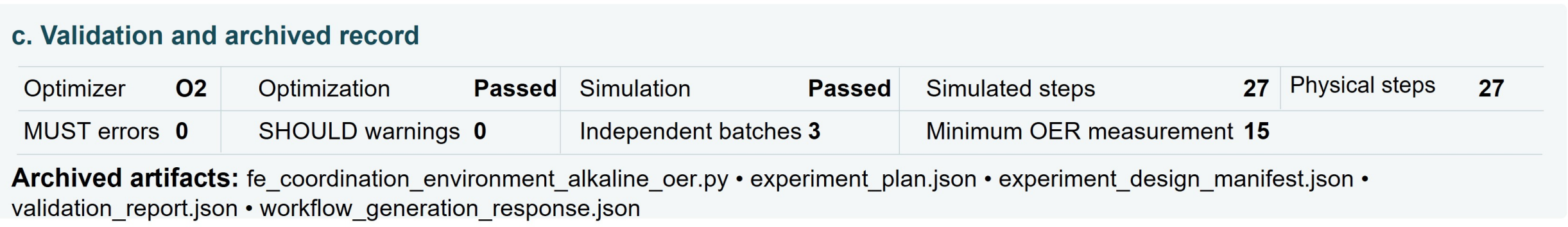

| Optimizer | O2 | Optimization | Passed | Simulation | Passed | Simulated steps | 27 | Physical steps | 27 |
|---|---|---|---|---|---|---|---|---|---|
| MUST errors | 0 | SHOULD warnings | 0 | Independent batches | 3 | Minimum OER measurement | 15 | | |

**Archived artifacts:** fe_coordination_environment_alkaline_oer.py • experiment_plan.json • experiment_design_manifest.json • validation_report.json • workflow_generation_response.json

Supplementary Fig. S7 | Full Fe coordination-environment comparison and alkaline OER workflow. The figure summarizes the experimental design used to investigate how the existing form and coordination environment of Fe influence the alkaline OER performance of NiFe-based catalysts. Five Fe-introduction routes are compared under matched synthesis and OER conditions, followed by a sequential workflow for preparation, characterization, and electrochemical testing. The archived validation results confirm successful workflow generation and simulation, while route-specific Fe coordination motifs are treated as working hypotheses requiring independent experimental verification.

## Full LDH–HMF synthesis, electrochemistry and multimodal analysis workflow

### a. Scientific intent

**Objective:** Design an integrated synthesis, electrochemical testing and multimodal characterization experiment for LDH catalysts toward alkaline 5-hydroxymethylfurfural (HMF) oxidation. The catalyst preparation and electrode-fabrication steps must follow a time-locked sequential workflow. After electrochemical testing, the solid electrode and liquid reaction products are required to undergo parallel contact-angle measurement, XRD analysis and liquid-phase product analysis, followed by integrated evaluation of the results.

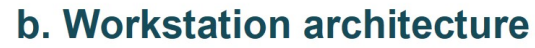

### b. Workstation architecture

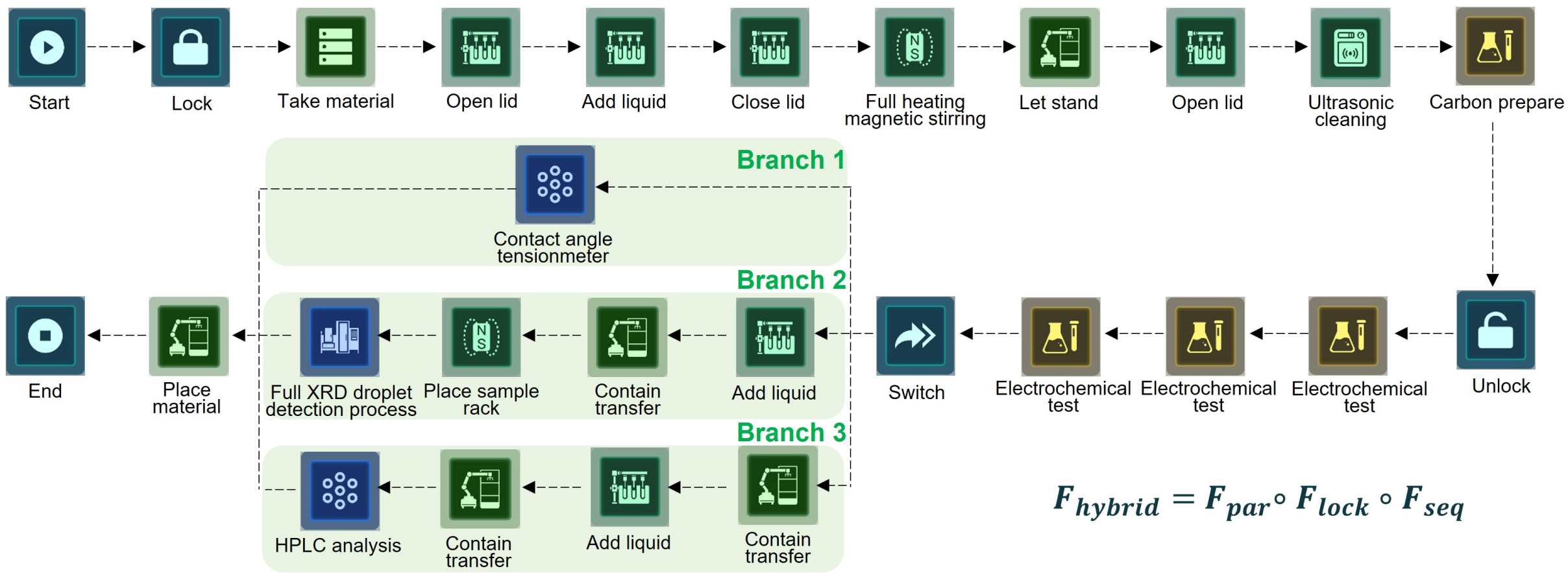

Join and integrated evaluation: Activity • selectivity • wettability • phase stability

### c. Validation and archived record

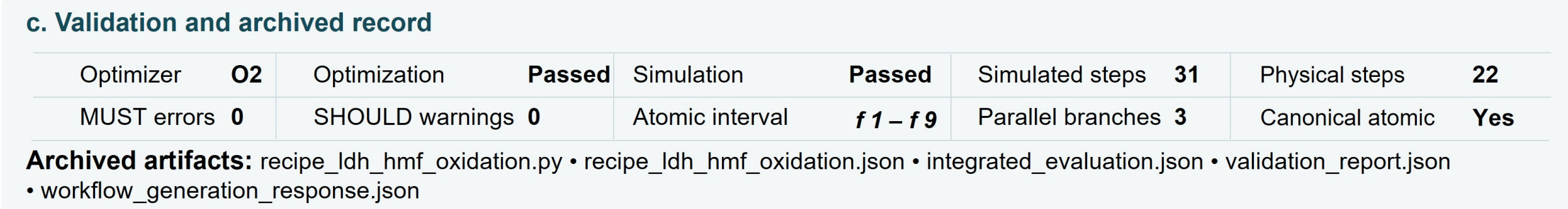

| | | | | | | | | | |
|---|---|---|---|---|---|---|---|---|---|
| Optimizer | O2 | Optimization | Passed | Simulation | Passed | Simulated steps | 31 | Physical steps | 22 |
| MUST errors | 0 | SHOULD warnings | 0 | Atomic interval | $f1-f9$ | Parallel branches | 3 | Canonical atomic | Yes |

**Archived artifacts:** recipe_ldh_hmf_oxidation.py • recipe_ldh_hmf_oxidation.json • integrated_evaluation.json • validation_report.json • workflow_generation_response.json

Supplementary Fig. S8 | Integrated synthesis, electrochemical testing, and multimodal characterization workflow for LDH catalysts toward alkaline HMF oxidation. The workflow combines a time-locked catalyst preparation sequence with sequential electrochemical evaluation, followed by parallel contact-angle, XRD, and liquid-phase product analyses. Results from all branches are subsequently joined for integrated evaluation, with archived validation confirming successful workflow generation and simulation.

## Furfural hydrogenation and bounded GC optimization workflow

### a. Scientific intent and analytical requirements

**Objective:** Design an automated catalyst synthesis and closed-loop analytical optimization experiment for the electrocatalytic hydrogenation of furfural to furfuryl alcohol. After catalyst preparation, a closed-loop algorithm is required to iteratively optimize the gas-chromatographic separation of furfural, furfuryl alcohol and major liquid by-products until predefined peak-resolution criteria are satisfied. The optimized method will then be applied to quantify the reaction mixture and evaluate furfural conversion, furfuryl alcohol yield and product selectivity. Need to use a loop node.

**Monitored analytes:** furfural • furfuryl alcohol • tetrahydrofurfuryl alcohol • cyclopentanone • cyclopentanol

**All pass criteria:** Rs (target pair) ≥ 1.50 • Rs (nearest major by product) ≥ 1.50 • max tailing ≤ 2.00 • retention time RSD ≤ 1.00%.

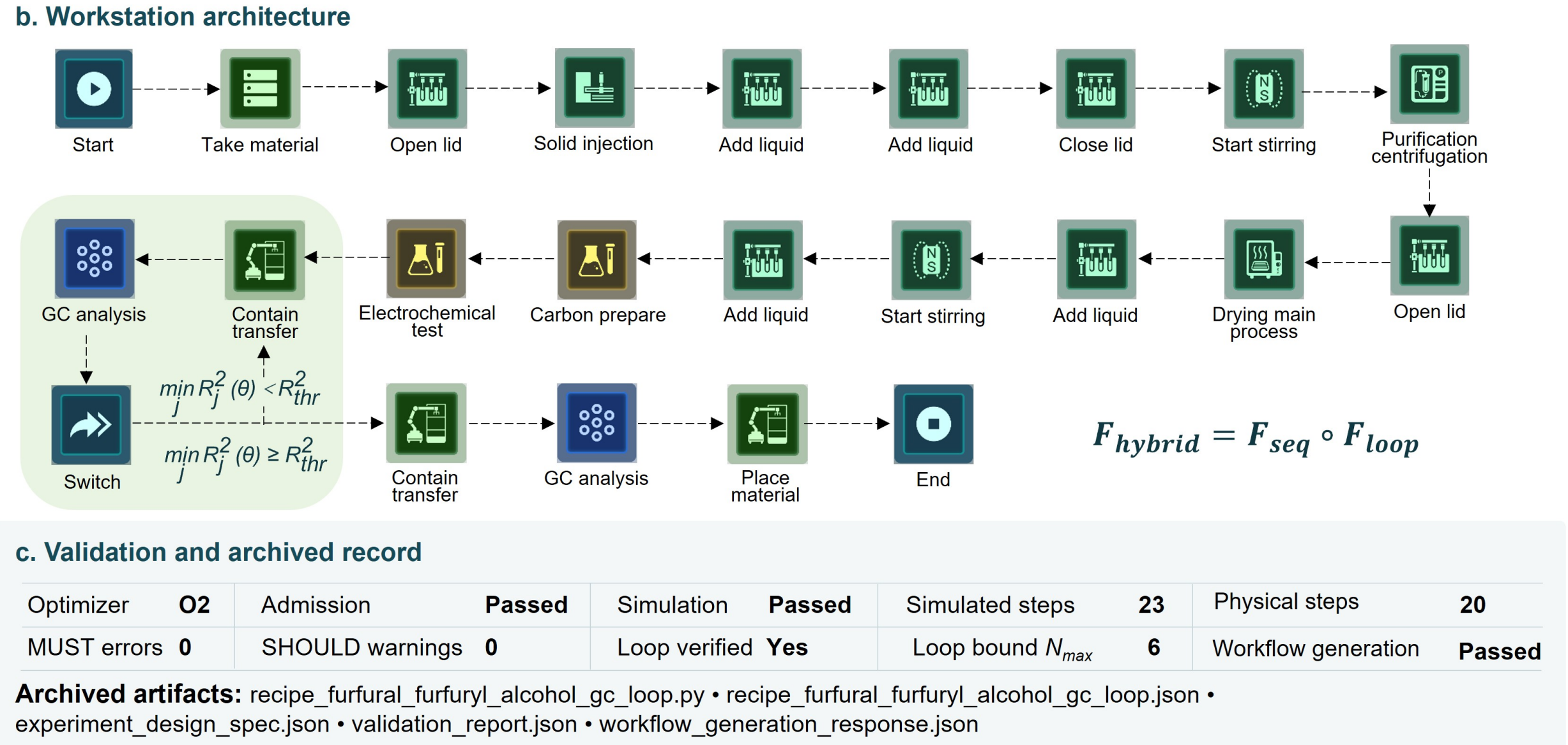

| Optimizer | O2 | Admission | Passed | Simulation | Passed | Simulated steps | 23 | Physical steps | 20 |
|---|---|---|---|---|---|---|---|---|---|
| MUST errors | 0 | SHOULD warnings | 0 | Loop verified | Yes | Loop bound $N_{max}$ | 6 | Workflow generation | Passed |

**Archived artifacts:** recipe_furfural_furfuryl_alcohol_gc_loop.py • recipe_furfural_furfuryl_alcohol_gc_loop.json • experiment_design_spec.json • validation_report.json • workflow_generation_response.json

Supplementary Fig. S9 | Automated furfural hydrogenation and closed-loop GC optimization workflow. The workflow integrates automated Cu-catalyst preparation, duplicate constant-potential electrocatalytic hydrogenation, and iterative GC method optimization for furfural, furfuryl alcohol, and major liquid by-products. GC conditions are iteratively refined until predefined separation criteria are satisfied or the maximum iteration limit is reached, after which the accepted method is used for quantitative analysis of conversion, yield, selectivity, and carbon balance. Archived validation results confirm successful workflow generation and simulation.

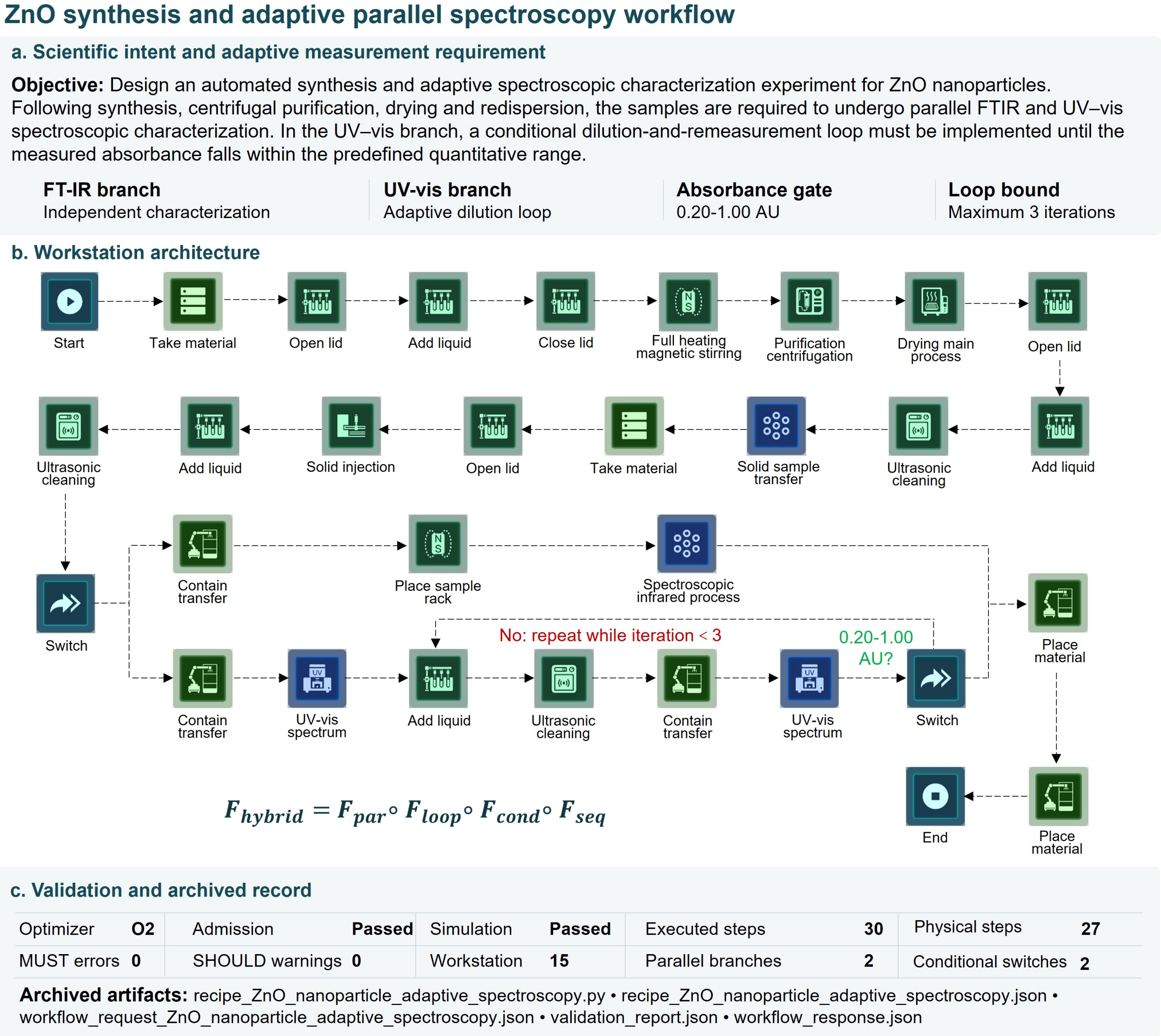

Supplementary Fig. S10 | Automated ZnO synthesis and adaptive parallel spectroscopy workflow. The workflow integrates ZnO nanoparticle synthesis, centrifugation, drying, and redispersion with parallel FTIR and UV–Vis characterization. The UV–Vis branch incorporates a bounded dilution-and-remeasurement loop until the predefined quantitative absorbance range is reached or the iteration limit is met. Archived validation confirms successful workflow generation and simulation.

**SI Guide entries**

**Supplementary Information**

**Supplementary Information. Supplementary Notes 1–12, Supplementary Tables S1–S12 and Supplementary Figs. S1–S10 supporting the formal representation, Function-Skill implementation, stateful simulation and four workflow cases.**

**Supplementary Data**

Supplementary Data 1. Capability and operation registry. Machine-readable inventory of the final workstation specifications, Function-Skill operations, accepted states, arguments, constraints, state changes, returns, source provenance, versions and checksums. File: Supplementary_Data/Supplementary_Data_1.xlsx

Supplementary Data 2. Formal grammar and canonical workflows. JSON definitions of the workflow grammar and canonical AST/JSON records for Fig. 1d and the four workflows in Extended Data Fig. 1a–d, with node-to-operation maps. File: Supplementary_Data/Supplementary_Data_2.json

Supplementary Data 3. Four-case evidence record. Exact prompts, recipe versions, workflow metrics, simulation results, structured diagnostics for Fe-coordination/OER, LDH–HMF, furfural and ZnO workflows. File: Supplementary_Data/Supplementary_Data_3.xlsx

Supplementary Data 4. Reproducibility manifest. Software, model and configuration versions, commands, environment information, filenames, SHA-256 checksums and links to author-supplied release artifacts. File: Supplementary_Data/Supplementary_Data_4.xlsx

**Author-supplied Additional Files**

**Additional File 1. Source-code archive.** Final optimizer, simulator, converter, configuration, dependency and reproduction files.

**Additional File 2. Function-Skill library.** Frozen 49-workstation capability library.

**Additional File 3. Extended Data Fig. 1 workflow archive.** Exact prompts, archived recipes, canonical JSON plans and validation reports for the four cases; physical-execution logs are not included. File: AF-03 Extended Data Fig. 1 workflow archive.zip (legacy filename retained).

**Additional File 4. Release-page record.** Author-supplied internal release-address record. File: AF-04 Release page.txt. T=33he recorded route is not a verified persistent public or reviewer-access repository.